\documentclass[11pt]{article}

\usepackage[
    backend=biber,
    style=nature,
    sorting=none
]{biblatex}
\usepackage{graphicx}
\usepackage{float}
\usepackage{placeins}
\usepackage{ragged2e}
\usepackage{caption}
\usepackage{microtype}
\usepackage{setspace}

\usepackage{newtxtext}
\usepackage{newtxmath}

\usepackage[
    a4paper,
    left=3.0cm,
    right=3.0cm,
    top=2.2cm,
    bottom=2.5cm
]{geometry}

\usepackage{hyperref}

\hypersetup{
    colorlinks=true,
    urlcolor=blue,
    linkcolor=black,
    citecolor=black
}
\newcommand{\naturefigurelegend}[3]{%
    \refstepcounter{figure}%
    \vspace{3pt}%
    \noindent
    \begin{minipage}{\textwidth}
        \RaggedRight
        \footnotesize
        \setstretch{0.96}
        \setlength{\parindent}{0pt}%
        \setlength{\parskip}{0pt}%
        \textbf{Fig. \thefigure\,$|$\,#2}\,
        #3
        \label{#1}%
    \end{minipage}
}

\newcommand{\naturefigure}[4]{%
    \begin{figure}[!htbp]
        \centering
        \includegraphics[
            width=\textwidth,
            height=0.53\textheight,
            keepaspectratio
        ]{#1}
        \par
        \naturefigurelegend{#2}{#3}{#4}
    \end{figure}
}

\newcommand{\naturefigurewide}[4]{%
    \begin{figure}[!htbp]
        \centering
        \includegraphics[
            width=\textwidth,
            keepaspectratio
        ]{#1}
        \par
        \naturefigurelegend{#2}{#3}{#4}
    \end{figure}
}

\usepackage{etoc}
\begin{document}

\begin{center}

% Title
% {\fontsize{18}{22}\selectfont\bfseries
% Knowledge-enhanced pretraining shifts the data-scaling curve of single-cell foundation models
% \par}

{\fontsize{18}{22}\selectfont\bfseries
Towards a knowledge-enhanced single-cell foundation model
\par}
\vspace{0.9em}

% Authors
% Final author names can be inserted here, for example:
% Authors
{\large
\mbox{Hanqing Zhang\textsuperscript{1,2,\ensuremath{\dagger}}},
\mbox{Jie Bao\textsuperscript{1,\ensuremath{\dagger}}},
\mbox{Mei Ma\textsuperscript{1}},
\mbox{Shuai Liu\textsuperscript{1}},
\mbox{Jiaying Ma\textsuperscript{1}},\\[0.25em]
\mbox{Jiaguan Liu\textsuperscript{1}},
\mbox{Jiaxiao Li\textsuperscript{1}},
\mbox{Zhenbo Li\textsuperscript{2}},
\mbox{Wenwen Gong\textsuperscript{2}},
\mbox{Zhijun Cao\textsuperscript{1,*}}
\par}

\vspace{0.7em}

% Affiliations
{\fontsize{8.5}{11}\selectfont
\textsuperscript{1}College of Animal Science and Technology,
China Agricultural University, Beijing 100193, China\\
\textsuperscript{2}College of Information and Electrical Engineering,
China Agricultural University, Beijing 100193, China
\par}

\vspace{0.4em}

% Equal contribution and correspondence
{\fontsize{8.5}{11}\selectfont
\textsuperscript{*}Corresponding author(s). Email(s):
\href{mailto:caozhijun@cau.edu.cn}{caozhijun@cau.edu.cn} \\
\textsuperscript{\ensuremath{\dagger}}These authors contributed equally to this work.

\par}

\end{center}

\vspace{1.5em}

\begin{abstract}
Single-cell foundation models (scFMs) increasingly rely on large-scale transcriptomic pretraining, yet expanding pretraining data can yield diminishing gains while substantially increasing computational cost. Our data scaling analyses showed that incorporating biological knowledge, including cell-level text annotation and gene-level regulatory information, provided additional scaling dimension than simply increasing data size. Motivated by this observation, we present scKITE, a simple yet effective scFM that integrates cell-annotation and gene-regulatory supervision into a shared transcriptomic Transformer encoder through lightweight auxiliary decoders. These decoders are used only during pretraining and subsequently discarded, yielding a general-purpose encoder enriched with biological knowledge for downstream applications. With only 179,067 pretraining samples, i.e., less than 0.5\% of those used by previous strong scFMs, scKITE outperformed these models across diverse downstream tasks, highlighting knowledge-enhanced pretraining as a promising paradigm for biologically grounded scFMs.

%%scKITE, kite风筝，靠外力飞上天，隐喻知识驱动的内涵。
%%

\end{abstract}

% =========================================================
% Introduction
% =========================================================

\section{Introduction}

Advances in single-cell RNA sequencing (scRNA-seq) have enabled high-resolution profiling of cellular heterogeneity across diverse tissues, developmental stages, disease states and experimental conditions
\cite{Macosko2015DropSeq,Zheng2017MassivelyParallel}. The resulting large-scale transcriptomic datasets have motivated the development of single-cell foundation models (scFMs), which learn transferable representations of genes and cells and provide a general framework for modeling cellular states from transcriptomic profiles \cite{Regev2017HumanCellAtlas}.

Scaling the size of pretraining corpora has become a major direction in the
development of scFMs. Geneformer established large-scale self-supervised
pretraining of single-cell transcriptomes for transferable prediction in
network biology, followed by models such as scGPT and scFoundation that
extended transcriptomic pretraining toward broader cell- and gene-level
applications \cite{theodorisTransferLearningEnables2023,Cui2024scGPT,
haoLargescaleFoundationModel2024,Yang2022scBERT,Wen2024CellPLM}. More recently, updated Geneformer models
have expanded the pretraining corpus to include more than 100 million
single-cell transcriptomes and increased model capacity to hundreds of
millions of parameters, further improving representation quality and
zero-shot prediction
\cite{Chen2026GeneformerScaling}. Together, these studies have demonstrated
that large-scale transcriptomic pretraining can capture substantial
transferable information from single-cell data
\cite{Heimberg2025SCimilarity,Rosen2026UCE,Gandhi2025TahoeX1}. However, continued scaling also introduces substantial computational and
practical costs. Pretraining transformer-based scFMs on increasingly large
corpora requires specialized hardware and extended training time, making
model development progressively more resource intensive. These costs become
particularly important when additional training data provide only limited
improvements in downstream performance. A recent systematic evaluation showed
that current scFMs can reach a performance plateau using only a fraction of the full pretraining corpus, with no clear data-scaling behavior comparable to that observed in large language models
\cite{DenAdel2026PretrainingScale,Kaplan2020ScalingLaws}. These observations
motivate complementary strategies beyond transcriptomic data scaling,
including enriching pretraining objectives with biologically meaningful
information.

Cellular systems are inherently complex, arising from coordinated interactions among genes and regulatory programs across diverse biological contexts~\cite{dupire2026fifteen}. This complexity has motivated the incorporation of biological knowledge into single-cell foundation models as an additional source of supervision for representation learning. Two complementary and readily accessible forms of knowledge are particularly relevant: natural-language descriptions that encode cell identity and biological context, and regulatory programs that describe transcription factor (TF)-centered regulatory relationships with their target genes within cellular states. Natural-language annotations provide information about cell identity, tissue context and functional state that may not be directly inferred from expression profiles alone, whereas regulatory programs describe the gene-regulatory relationships that shape cellular states. Incorporating these complementary sources of cell- and gene-level information may therefore help scFMs learn representations that better capture both cellular context and underlying regulatory structure. Recent studies have shown that transcriptomic profiles can be linked to biological text descriptions ~\cite{changTracingRiseBiomedical2026,Schaefer2025CellWhisperer,Zhao2024LangCell,Levine2024Cell2Sentence}. Moreover, GRN (Gene Regulatory Networks) inference tools such as SCENIC and pySCENIC also further provide feasible approaches for inferring regulatory networks at single-cell resolution from transcriptomic data ~\cite{aibarSCENICSinglecellRegulatory2017,vandesandeScalableSCENICWorkflow2020,BravoGonzalezBlas2023SCENICPlus}.  These two knowledge sources potentially offer complementary information for learning biologically informative transcriptomic representations.

Motivated by this knowledge-enhanced perspective, we systematically characterized data scaling by varying the pretraining corpus size while holding model architecture and evaluation settings fixed. We found that performance gains from increasing data rapidly diminished, whereas knowledge-enhanced pretraining consistently improved performance across data scales, achieving an average 28.8\% relative improvement in downstream task performance compared with scKITE without knowledge-enhanced training. These results suggest that biological knowledge can shift the data-scaling curve, reducing reliance on continued expansion of the pretraining corpus. Guided by this observation, we developed scKITE (\textbf{s}ingle-\textbf{c}ell \textbf{K}nowledge-\textbf{I}ntegrated \textbf{T}ransform\textbf{e}r), a knowledge-enhanced single-cell foundation model that incorporates annotation and regulatory supervision into transcriptomic pretraining. scKITE employs two lightweight auxiliary decoders for annotation and regulon prediction, providing complementary supervision on cellular identity and gene regulation and thereby enriching the shared transcriptomic encoder with biological knowledge. After pretraining, both decoders are discarded, leaving a single knowledge-enhanced encoder that serves as a general-purpose representation model for downstream cell- and gene-level applications.

We demonstrate that knowledge-enhanced pretraining improves transcriptomic
representations and downstream performance in scKITE. Using only 179,067 ($\sim$ 0.18M) pretraining samples, less than 0.5\% of the data used by representative scFMs, scKITE outperformed existing SOTA scFMs, including Geneformer ($\sim$ 30M pretraining samples), scGPT ($\sim$ 33M pretraining samples) and scFoundation ($\sim$ 50 M pretraining samples), across diverse downstream tasks, and GEARS, a classical perturbation prediction model. Compared with the corresponding scKITE without knowledge enhancement, scKITE achieved substantial improvements at the cell level, including a 64.0\% increase in mean zero-shot macro-F1 across five cell type annotation benchmarks and a 22.8\% increase in mean batch-integration overall score across four datasets. At the gene level, scKITE reduced Top-20 DE MSE by 19.8\% across perturbation-response benchmarks.

In summary, our results establish knowledge-enhanced pretraining as an effective strategy for learning more biologically grounded single-cell representations while reducing reliance on continued data scaling. By effectively integrating two structurally distinct forms of biological knowledge (i.e., cell-level textual annotations and gene-level regulatory relationships) into a shared transcriptomic representation, scKITE achieves strong downstream performance with substantially fewer pretraining samples, highlighting a broader route for advancing scFMs through biologically grounded learning rather than data scale alone.

\section{Results}

% =========================================================
% Result 1 / Figure 1
% =========================================================

\subsection{Knowledge-enhanced pretraining improves transcriptomic representations beyond data scaling}

scKITE integrates complementary biological knowledge into a shared transcriptomic representation through a two-stage pretraining framework, achieving greater performance than simply scaling the amount of pretraining data.

We demonstrate that performance gains from increasing data rapidly diminished, whereas knowledge-enhanced supervision consistently improved performance across data scales. The pretraining corpus was divided into 358,134 training profiles and a fixed validation set of 18,849 profiles. Across data fractions, scKITE consistently maintained stronger normalized downstream performance than scKITE (w/o knowledge enhancement) across cell type annotation, batch integration and perturbation-response prediction (Fig.~\ref{fig:framework}a). Notably, scKITE (w/o knowledge enhancement) reached a performance plateau after scaling to 50\% of the pretraining data, showing limited additional gains from further corpus expansion. However, across four pretraining scales, scKITE achieved an average 28.8\% relative improvement in downstream task performance compared with scKITE (w/o knowledge enhancement). We use 179,067 training profiles for scKITE training and performance evaluation in all subsequent experiments.

To provide complementary biological supervision, each transcriptomic profile was paired with annotation and regulatory knowledge (Fig.~\ref{fig:framework}b). The pretraining corpus was derived from the CELLxGENE~\cite{Chan2024CELLxGENE} component of CellWhisperer~\cite{Schaefer2025CellWhisperer}, comprising pseudo-bulk transcriptomic profiles paired with metadata-derived natural-language descriptions of cell identity and biological context. These descriptions were tokenized into annotation knowledge sequences for language-based supervision. In parallel, pySCENIC~\cite{vandesandeScalableSCENICWorkflow2020} was used to infer regulatory activity from each transcriptomic profile (see Supplementary Methods~\ref{sec:regulon_supervision}). Activated regulons were identified, and a subset was sampled and serialized into sequences encoding transcription factor–target relationships. Each transcriptomic profile was therefore associated with two complementary supervision targets: an annotation sequence capturing cellular identity and context, and a regulon sequence capturing underlying regulatory programs.

Integrating two structurally distinct forms of biological knowledge into scFMs is non-trivial. scKITE addresses this challenge through a two-stage pretraining framework comprising transcriptomic self-supervised pretraining followed by knowledge-enhanced pretraining (Fig.~\ref{fig:framework}c).  In Stage~1, a Transformer-based encoder is pretrained using a masked-expression reconstruction objective. Stage~2 is initialized from the Stage~1 checkpoint and retains the expression reconstruction objective while introducing annotation and regulon prediction as additional knowledge-enhanced supervision. Two lightweight auxiliary decoders, each connected to the shared encoder through cross-attention, simultaneously predict annotation and regulon knowledge sequences, thereby incorporating complementary biological knowledge into the shared transcriptomic representation. Specifically, Stage~1 is optimized using the masked-expression reconstruction
loss ($\mathcal{L}_{\mathrm{expr}}$), whereas Stage~2 jointly optimizes the
shared encoder using $\mathcal{L}_{\mathrm{expr}}$ together with
autoregressive cross-entropy losses for annotation sequence generation
($\mathcal{L}_{\mathrm{ann}}$) and regulon sequence generation
($\mathcal{L}_{\mathrm{reg}}$) (see Supplementary Methods~\ref{sec:stage2_pretraining}). Both auxiliary decoders access the full sequence of encoder hidden states through cross-attention and are used only during pretraining. After pretraining, the decoders are discarded, leaving a knowledge-enhanced encoder that can be directly transferred to downstream tasks without requiring annotation or regulon information at inference time. This design preserves a general-purpose transcriptomic representation while enriching it with complementary biological knowledge, enabling the encoder alone to serve as a transferable representation across downstream applications.

The retained scKITE encoder provides reusable representations at both cell
and gene levels (Fig.~\ref{fig:framework}d). Cell embeddings support cell type
classification, batch integration and cell organization analyses, whereas
gene embeddings support perturbation-response prediction, marker-gene analysis
and TF-regulon interpretation. Thus, complementary biological knowledge introduced during pretraining is consolidated into a single reusable encoder, providing a unified representation for downstream analyses across both cell- and gene-level tasks.

\naturefigurewide
{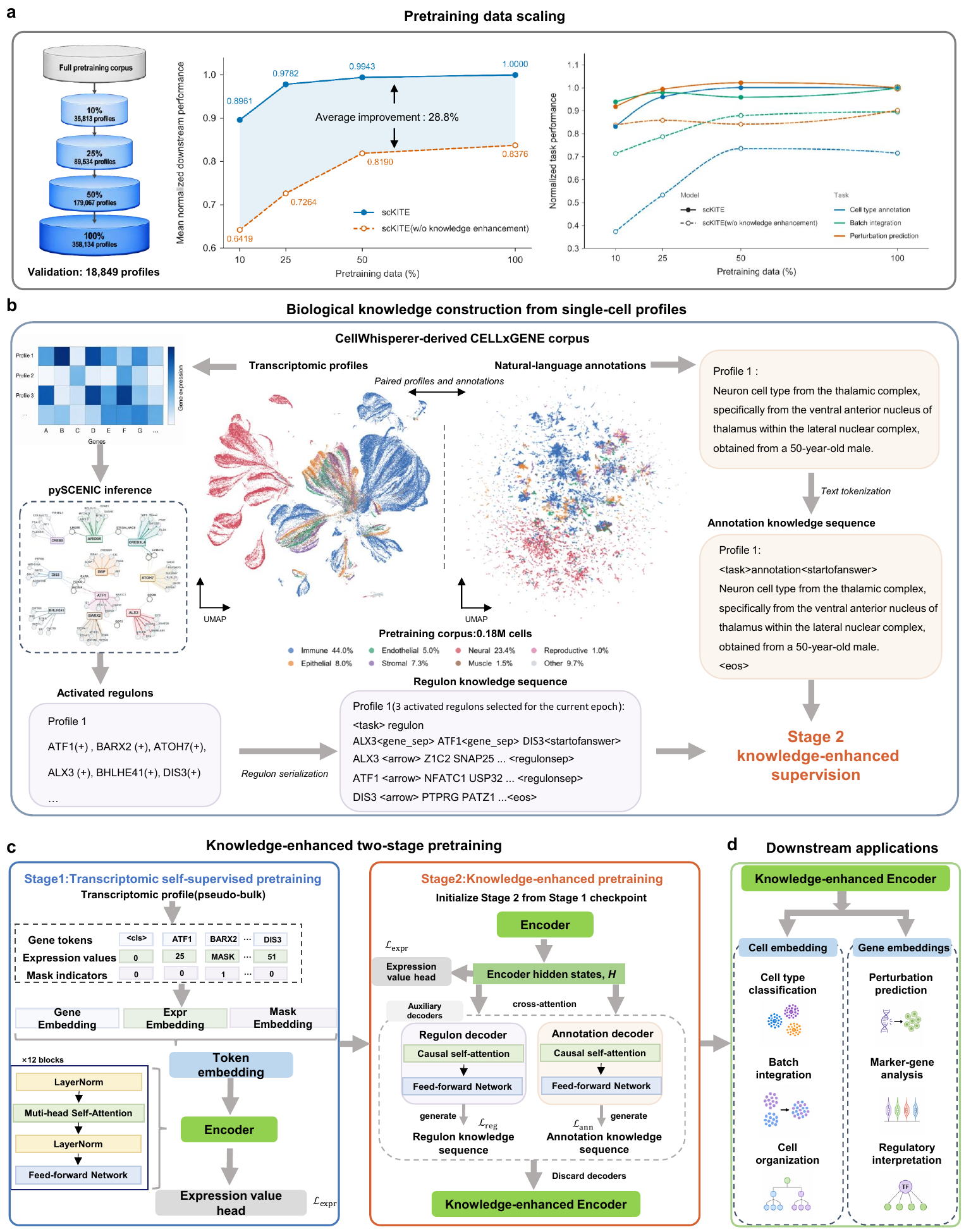}
{fig:framework}
{Knowledge-enhanced pretraining strategy of scKITE enables biologically informed single-cell representations.}
{\textbf{a,}
Effect of pretraining corpus size and biological knowledge enhancement on scKITE performance. Models pretrained using different fractions of the corpus are evaluated across cell-type annotation, batch integration and perturbation-response prediction tasks.
\textbf{b,}
Construction of complementary biological supervision from the CellWhisperer-processed CELLxGENE corpus. Transcriptomic profiles are paired with natural-language cell annotations and activated regulon programs to generate annotation and regulatory knowledge sequences for Stage 2 knowledge-enhanced pretraining.
\textbf{c,}
Two-stage pretraining strategy of scKITE. Stage 1 performs transcriptomic self-supervised learning through masked expression modeling. Stage 2 initializes the encoder from Stage 1 and introduces auxiliary annotation and regulon decoders to inject biological knowledge into representation learning. The auxiliary decoders are removed after pretraining, and the knowledge-enhanced encoder is used for downstream tasks.
\textbf{d,}
Downstream applications of the pretrained scKITE encoder. Cell-level representations support cell-type annotation, batch integration and cell organization analysis, whereas contextual gene representations enable perturbation-response prediction and regulatory interpretation.
}
% =========================================================
% Result 2 / Figure 2
% =========================================================

\subsection{Knowledge-enhanced pretraining improves cell identity representation}

scKITE improves cell type annotation while better capturing hierarchical
immune organization and cross-tissue cell type relationships. We evaluated scKITE across
five benchmarks covering standard settings (Human Pancreas and Tabula
Sapiens), cross-disease generalization (Multiple Sclerosis), cross-cancer
generalization (Tumor-infiltrating Myeloid), and cross-tissue generalization
(Cross-tissue Immune Cell Atlas). scKITE was compared with scKITE (w/o knowledge enhancement) and representative single-cell foundation models, scGPT
and Geneformer, under zero-shot and full-fine-tuning settings. For the zero-shot cell type annotation task, frozen encoder representations
were directly used with a nearest-neighbor classifier for cell type
assignment. For the full-fine-tuning cell type annotation task, a
task-specific neural network classifier was optimized using the pretrained
representations as input(see Supplementary Methods~\ref{sec:cell_annotation}).

In the zero-shot setting, scKITE achieved the best or comparable performance across most evaluated datasets and metrics. Across the five benchmarks, scKITE
achieved a mean macro-F1 of 0.500, compared with 0.430 for scGPT and 0.399
for Geneformer. This advantage was largely retained after full fine-tuning, where scKITE
achieved the best performance across nearly all evaluated datasets and
annotation metrics, with only one metric slightly below scGPT. The mean
macro-F1 reached 0.668, compared with 0.561 for scGPT and 0.543 for
Geneformer (Fig.~\ref{fig:cell_classification}a). The zero-shot cell
embeddings of scKITE in the Tabula Sapiens and Cross-tissue Immune Cell Atlas datasets further illustrated the
organization of cell representations (Fig.~\ref{fig:cell_classification}b). The confusion matrix showed strong agreement
between predicted and annotated labels across most of the 20 common cell
types In the Tabula Sapiens dataset  (Fig.~\ref{fig:cell_classification}c).

Knowledge-enhanced pretraining enabled scKITE representations to better
capture the hierarchical organization of related immune cell compartments. The Cross-tissue Immune Cell Atlas organizes immune populations into three major compartments,
including myeloid cells, T and innate lymphoid cells, and B cells, which
represent major functional and developmental branches of the immune system \cite{DominguezConde2022}. Hierarchical clustering of cell-type centroids revealed clearer separation among the annotated myeloid, T and innate lymphoid, and B-cell compartments in the scKITE embedding space than in the scKITE (w/o knowledge enhancement) embedding space (Fig.~\ref{fig:cell_classification}d). Agreement between the resulting clusters and the annotated immune compartments increased from an ARI of 0.20 to 0.46 and from an NMI of 0.34 to 0.68. At the individual-cell level, pairwise cosine distances further revealed the hierarchical organization of immune cell relationships (Fig.~\ref{fig:cell_classification}e). Cells from the same cell type showed
the smallest distances, followed by different cell types within the same
immune compartment, whereas cells from different compartments showed the
largest distances. This separation among the three levels was more pronounced
in scKITE embeddings than in the scKITE (w/o knowledge enhancement) representation, indicating that
scKITE better captured the hierarchical organization reflected by annotated immune compartments among immune cell types.

Cross-tissue retrieval demonstrated that knowledge-enhanced
pretraining improved the ability of scKITE representations to identify cell types across different tissues. Across nine lymphoid and non-lymphoid tissues, cell type centroids were
matched across tissues within the same immune compartment, with retrieval
accuracy defined by whether the nearest target centroid corresponded to the
same annotated cell type. Across directional cross-tissue comparisons, scKITE consistently achieved higher Top-1 retrieval accuracy than the scKITE (w/o knowledge enhancement), increasing the
overall retrieval accuracy from 42.8\% to 66.2\%
(Fig.~\ref{fig:cell_classification}f). These results show that scKITE more
effectively preserves cell identity across distinct tissues.

Together, these results demonstrate that knowledge-enhanced pretraining
improves the quality of scKITE cell representations, enabling accurate cell
type annotation while preserving hierarchical immune organization and
cross-tissue cell type relationships.

\naturefigure
{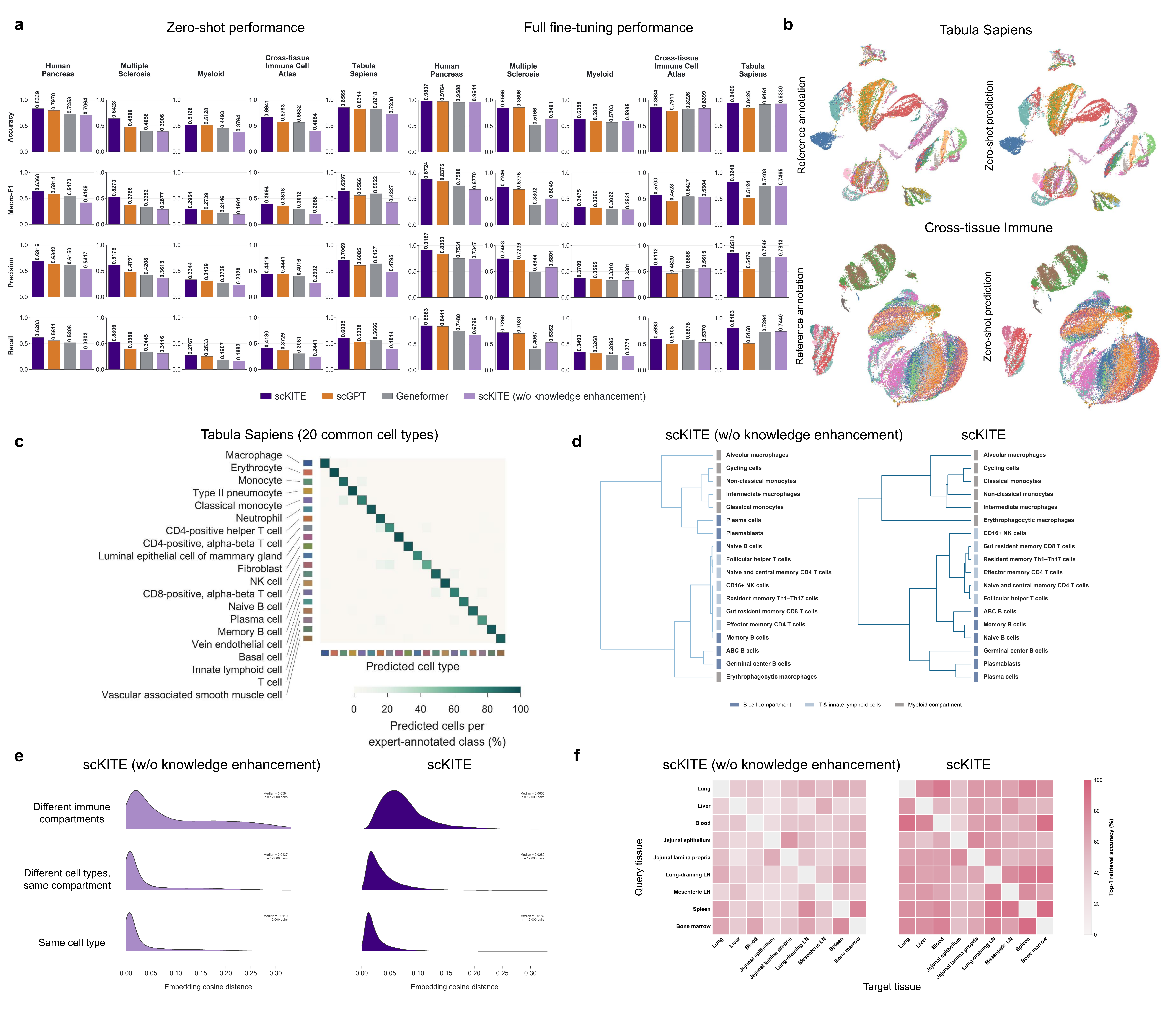}
{fig:cell_classification}
{Evaluation of scKITE representations for cell type annotation and biological organization.}
{
\textbf{a}, Zero-shot and fine-tuned cell-type annotation performance of scKITE, scKITE (w/o knowledge enhancement), scGPT and Geneformer across five benchmark datasets, including Human Pancreas, Multiple Sclerosis, Tumor-infiltrating Myeloid, Cross-tissue Immune Cell Atlas and Tabula Sapiens.
\textbf{b}, UMAP visualization of scKITE cell embeddings from the Tabula Sapiens and Cross-tissue Immune Cell Atlas datasets, colored by reference cell types and zero-shot predicted labels.
\textbf{c}, Confusion matrix comparing reference annotations and zero-shot predictions for 20 common cell types in the Tabula Sapiens dataset.
\textbf{d}, Hierarchical organization of immune cell types captured by scKITE and scKITE (w/o knowledge enhancement). Dendrograms show clustering of cell-type centroids colored by major immune compartments.
\textbf{e}, Pairwise cosine-distance distributions in scKITE and scKITE (w/o knowledge enhancement) embeddings for cells across different immune compartments, within the same compartment, and within the same cell type.
\textbf{f}, Cross-tissue retrieval accuracy of scKITE representations across nine lymphoid and non-lymphoid tissues in the Cross-tissue Immune Cell Atlas. Heatmaps show directional Top-1 retrieval accuracy between tissue pairs.
}

% =========================================================
% Result 3 / Figure 3
% =========================================================

\subsection{Knowledge-enhanced pretraining improves batch integration}

scKITE improves batch integration while preserving biologically meaningful
cell type structure across diverse scRNA-seq datasets. We examined four
integration benchmarks: Perirhinal cortex (2 batches), COVID-19 (18 batches),
Renal (3 batches), and Liver (16 batches). In our benchmarking experiments,
we compared scKITE with scKITE (w/o knowledge enhancement) and two representative
single-cell foundation models, scGPT and Geneformer. Integration performance was evaluated from two complementary aspects:
biological conservation and batch mixing. The AvgBIO score aggregates three
cell-type preservation metrics, normalized mutual information (NMI),
adjusted Rand index (ARI) and cell-type average silhouette width
(ASW$_{\mathrm{cell}}$), whereas the AvgBATCH score summarizes batch-mixing
metrics, including batch average silhouette width (ASW$_{\mathrm{batch}}$)
and graph connectivity.

Across four batch-integration benchmarks, scKITE consistently outperformed scKITE (w/o knowledge enhancement) and achieved competitive performance compared with existing scFMs
(Fig.~\ref{fig:batch_integration}a). Relative to scKITE (w/o knowledge enhancement), mean AvgBIO increased from approximately 0.338 to 0.474 and mean
AvgBATCH from approximately 0.836 to 0.938. scKITE also exceeded the strongest external baseline, Geneformer, by 6.5\% in AvgBIO and 4.2\% in AvgBATCH.
Together, these results indicate that scKITE achieves a better balance between biological conservation and batch-effect removal. Metric decomposition further showed broad improvements over scKITE (w/o knowledge enhancement) across the major integration metrics
(Fig.~\ref{fig:batch_integration}b). Averaged across the four datasets,
scKITE increased NMI by 0.179, ARI by 0.112 and cell-type ASW by 0.118.
Graph connectivity increased by 0.134 on average, and batch ASW increased by
0.071. Integration performance was further evaluated within individual cell types.
In the representative Perirhinal cortex and Liver datasets, scKITE showed
consistently high batch ASW and graph connectivity across most evaluated cell
types (Fig.~\ref{fig:batch_integration}c). These results indicate that scKITE
maintains effective batch mixing and local connectivity across diverse cell
types. Representative scKITE embeddings of the Perirhinal cortex and Liver datasets further showed that cells from different batches were well mixed
within the same cell types while distinct cell type structures remained
preserved (Fig.~\ref{fig:batch_integration}d,e). Overall, scKITE improves batch integration by improving batch mixing
while preserving biologically meaningful cell type organization across diverse
single-cell datasets.

\naturefigurewide
{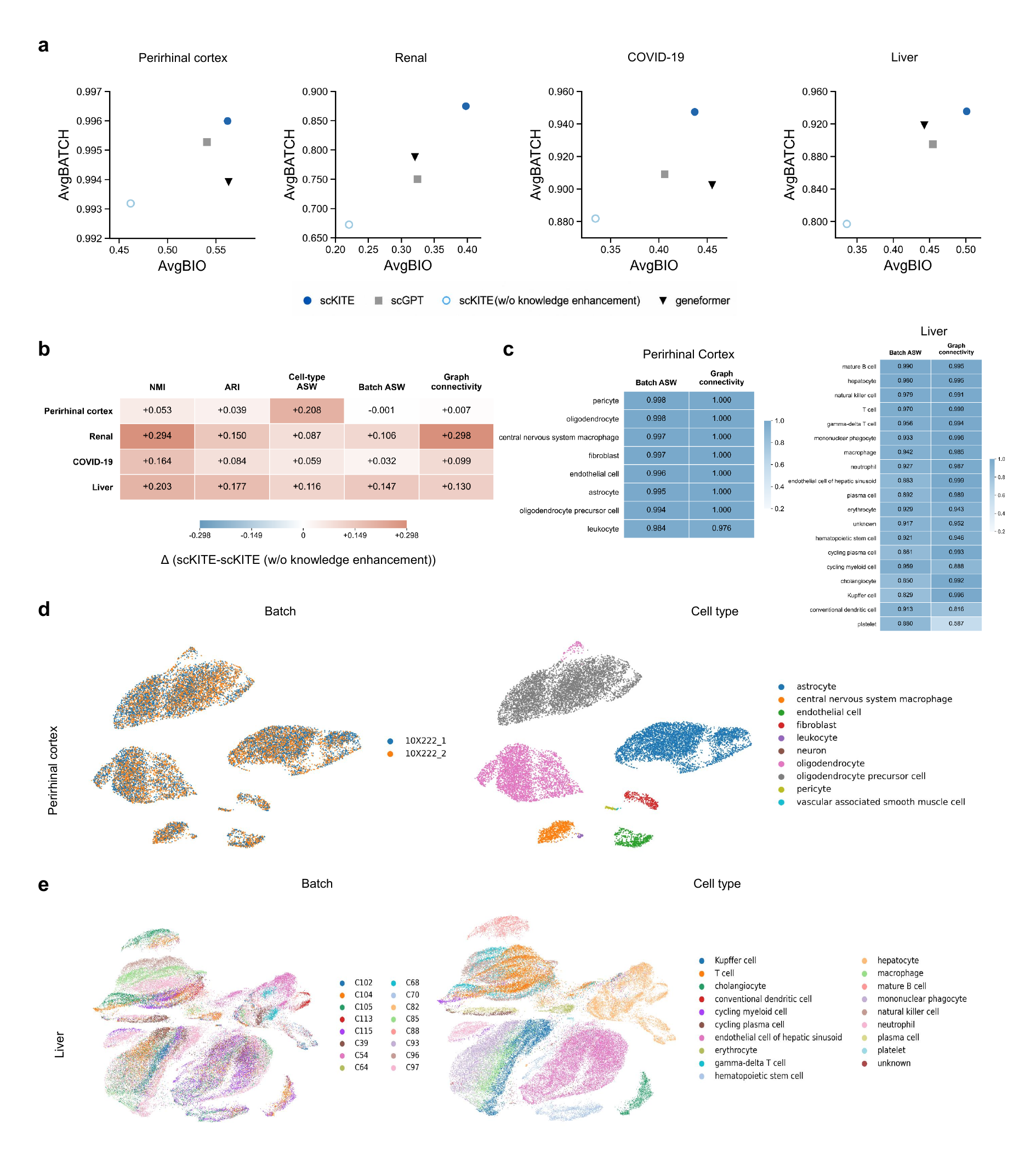}
{fig:batch_integration}
{Evaluation of scKITE representations for cross-dataset batch integration.}
{
\textbf{a}, Benchmarking of AvgBIO and AvgBATCH scores for scKITE,
scKITE (w/o knowledge enhancement), scGPT and Geneformer across the
Perirhinal cortex, Renal, COVID-19 and Liver datasets.
\textbf{b}, Improvements of scKITE over scKITE (w/o knowledge enhancement)
across integration metrics, including NMI, ARI, cell-type ASW, batch ASW and
graph connectivity. Values represent the difference between scKITE and the
knowledge-ablation model.
\textbf{c}, Cell-type-resolved evaluation of batch mixing for scKITE using
batch ASW and graph connectivity across individual cell types in the
Perirhinal cortex and Liver datasets.
\textbf{d}, UMAP visualization of scKITE embeddings for the Perirhinal cortex
dataset colored by cell type (right) and assay batch (left).
\textbf{e}, UMAP visualization of scKITE embeddings for the Liver dataset
colored by cell type (right) and donor batch (left).
}
% =========================================================
% Result 4 / Figure 4
% =========================================================

\subsection{Knowledge-enhanced pretraining improves gene context embeddings for perturbation-response prediction}

We tested whether the knowledge-enhanced gene context embeddings learned by scKITE improve genetic perturbation-response prediction using two Perturb-seq datasets, Norman\cite{Norman2019} and Replogle\cite{Replogle2022}. Gene context embeddings obtained from scKITE, scGPT and scFoundation were incorporated into a shared GEARS-based prediction framework (Fig.~\ref{fig:perturbation}a). GEARS combines gene embeddings, perturbation embeddings, gene co-expression graph and Gene Ontology graph to predict post-perturbation gene expression. The original GEARS model with native gene embeddings was included as a baseline.

scKITE gene context embeddings improved perturbation-response prediction performance across perturbation-response datasets and most perturbation generalization settings. scKITE achieved the lowest
overall Top-20 DE MSE in both the Norman and Replogle datasets
(Fig.~\ref{fig:perturbation}b), outperforming scKITE (w/o knowledge enhancement), GEARS and the evaluated single-cell foundation-model representations. We next examined prediction performance across different perturbation generalization settings in the Norman dataset. MSE was calculated over the
perturbation-specific top 20 non-dropout DE genes for Seen~0, Seen~1,
Seen~2 and unseen-single conditions. scKITE achieved the lowest MSE in
Seen~0, Seen~1 and Seen~2 (Fig.~\ref{fig:perturbation}c). Under the unseen-single setting, scKITE achieved lower MSE than GEARS, scGPT and scKITE (w/o knowledge enhancement), while scFoundation showed a lower MSE. Prediction performance was further assessed using Pearson correlation between
predicted and observed condition-mean expression profiles over the top
20 DE genes. scKITE achieved the highest correlation in Seen~1, Seen~2 and
unseen-single conditions, while showing a slightly lower correlation than
scGPT in the Seen~0 setting (Fig.~\ref{fig:perturbation}c). Across all four
settings, scKITE outperformed both the original GEARS baseline and scKITE (w/o knowledge enhancement). 

Prediction performance was further examined at the individual perturbation level in the Replogle dataset. scKITE achieved lower Top-20 DE MSE than GEARS across all evaluated perturbation conditions, with improvement observed in all 25 perturbations (25/25; Fig.~\ref{fig:perturbation}d), demonstrating consistent improvement across evaluated perturbation conditions. To further evaluate whether scKITE could recover biologically meaningful perturbation-responsive genes, we quantified the direction-correct recovery of top-20 DE genes in the Norman dataset. Compared with scKITE (w/o knowledge enhancement), scKITE achieved a higher recovery score across 97 held-out perturbation conditions, with a significant improvement measured by a paired Wilcoxon signed-rank test (P = 0.010; Fig.~\ref{fig:perturbation}e). These results indicate that knowledge enhancement improves the recovery of perturbation-responsive genes and their expression-change directions. To illustrate the predicted perturbation responses, we
visualized the representative
\textit{UBASH3B}+\textit{PTPN12} combinatorial perturbation in the Norman
dataset (Fig.~\ref{fig:perturbation}f), where scKITE reproduced the observed expression-change patterns across the top 20 DE genes. 

Finally, we evaluated whether scKITE could recover genetic interaction strength from predicted perturbation responses. Genetic interaction magnitudes were calculated following the GEARS interaction-analysis framework, and the agreement between predicted and ground-truth magnitudes was assessed using Pearson correlation. scKITE achieved a higher correlation than GEARS (Pearson \(r=0.337\) versus \(0.229\); Fig.~\ref{fig:perturbation}g), indicating improved recovery of relative genetic interaction strength for double-gene perturbations. Together, these results demonstrate that knowledge-enhanced gene context
embeddings provide transferable representations for predicting transcriptional
responses across diverse genetic perturbation settings.

\naturefigure
{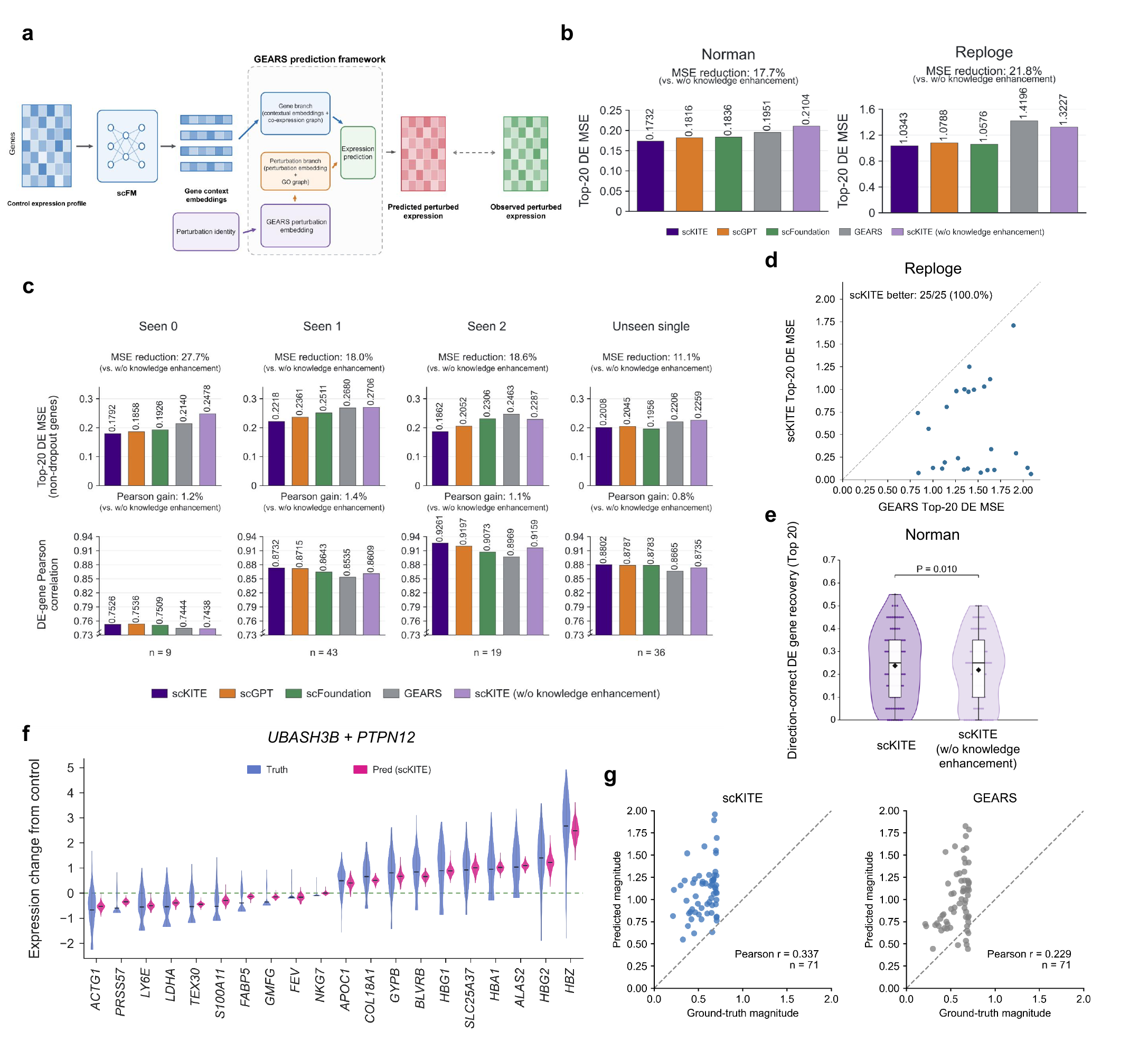}
{fig:perturbation}
{Evaluation of gene context embeddings for genetic perturbation-response prediction.}
{
\textbf{a}, Overview of the GEARS-based framework using contextual gene
embeddings learned by scKITE for perturbation-response prediction. The
pretrained gene embeddings are combined with perturbation information to
predict perturbation responses.
\textbf{b}, Top-20 DE gene mean squared error (MSE) between predicted and observed expression profiles across the Norman and Replogle datasets. Performance is compared across scKITE, scGPT, scFoundation, GEARS and scKITE (w/o knowledge enhancement).
\textbf{c}, Perturbation-response prediction performance under different
generalization settings in the Norman dataset. Seen0, Seen1 and Seen2
represent increasing levels of perturbation overlap between training and
evaluation, whereas unseen single represents unseen single-gene
perturbations.
\textbf{d}, Perturbation-level comparison of Top-20 DE gene MSE between
scKITE and GEARS across evaluated perturbation conditions in the Replogle
dataset. Each point represents one perturbation condition.
\textbf{e}, Direction-correct recovery of Top-20 DE genes in the Norman dataset.
Each point represents one unique perturbation condition in the Norman test set after removing equivalent condition aliases (\(n=97\)). Violin plots show
DE gene recovery score distributions across perturbations, with boxes
indicating the interquartile range and black diamonds indicating the mean.
scKITE was compared with scKITE (w/o knowledge enhancement) using a
two-sided paired Wilcoxon signed-rank test.
\textbf{f}, Comparison of predicted and observed expression-change
distributions for Top-20 DE genes under the
\textit{UBASH3B}+\textit{PTPN12} combinatorial perturbation. Violin plots
show cell-level expression-change distributions, with horizontal lines
indicating mean values.
\textbf{g}, Recovery of genetic interaction magnitude for held-out
double-gene perturbations. Each point represents one double-gene
perturbation (\(n=71\)). Scatter plots compare predicted and ground-truth
interaction magnitudes for scKITE and GEARS, with dashed lines indicating
perfect agreement (\(y=x\)). Pearson correlations are shown for each model.
}
% =========================================================
% Result 5 / Figure 5
% =========================================================

\subsection{Biological interpretation of knowledge-enhanced representations}

The improved performance of scKITE across cell- and gene-level benchmarks
highlights the value of incorporating biological supervision into single-cell
pretraining. We therefore investigated how annotation and regulatory
supervision were reflected in the shared encoder representation through
decoder-to-encoder cross-attention analysis. During Stage~2, both decoders
access encoder hidden states through cross-attention, enabling attention
patterns to provide an interpretable view of the gene states preferentially
accessed during biological target reconstruction (Fig.~\ref{fig:regulatory_mechanism}a).

Annotation supervision was assessed by examining whether decoder attention
preferentially focused on cell-type marker genes. Marker genes for each of the
198 cell types were identified from the training set using one-versus-rest
differential-expression analysis with a Wilcoxon rank-sum test
\cite{stuart2019comprehensive,wilcoxon1945individualcomparisons}. These training-derived marker genes
were then evaluated in the held-out validation set by comparing their
decoder-to-encoder cross-attention percentiles with matched background genes. Attention percentiles were used to quantify the relative ranking of genes
based on decoder cross-attention scores (see Supplementary Methods~\ref{sec:stage2_pretraining}). Across 198
evaluable cell types, marker genes showed higher attention percentiles than
matched background genes in 196 cell types, resulting in strong paired
enrichment across cell types ($P=3.7\times10^{-34}$;
Fig.~\ref{fig:regulatory_mechanism}b). To further evaluate whether attention rankings recovered established cell identity
signals, canonical marker genes for 13 representative cell types were curated
from CellMarker 2.0\cite{Hu2023CellMarker2}. Representative cell types illustrated that decoder attention preferentially
focused on established cell identity markers. In NK cells,
markers including \textit{GNLY}, \textit{NKG7}, \textit{KLRD1} and
\textit{PRF1} exhibited high median attention percentiles, whereas neutrophil
markers including \textit{S100A8}, \textit{S100A9}, \textit{FCGR3B},
\textit{CSF3R} and \textit{CXCR2} showed similar enrichment patterns
(Fig.~\ref{fig:regulatory_mechanism}c). Across 61 evaluable canonical marker
genes from 13 cell types, 22 markers (36.1\%) were recovered within the top 10
attention-ranked genes, increasing to 27 (44.3\%) and 36 (59.0\%) within the
top 15 and top 25 genes, respectively (Fig.~\ref{fig:regulatory_mechanism}d).
Together, these results indicate that annotation supervision directs decoder
attention toward cell-type marker genes.

Regulatory supervision was assessed by examining whether decoder attention
preferentially accessed genes within externally supported TF regulatory
programs. We used CollecTRI\cite{MullerDott2023CollecTRI} to annotate
TF-target relationships for 45 active TF programs and compared supported target
genes with matched non-target genes using regulon decoder cross-attention.
Across the 45 evaluable TF programs, supported targets showed higher attention
than matched non-target genes for 32 TFs, indicating a tendency toward
preferential attention to externally supported regulatory targets
($P=5.9\times10^{-2}$; Fig.~\ref{fig:regulatory_mechanism}e). TF-level analysis revealed that regulatory attention patterns varied across
TF programs. Several TF programs showed higher attention toward CollecTRI-
supported target genes than matched non-target genes (Fig.~\ref{fig:regulatory_mechanism}f). In the representative \textit{TBX21} regulatory program, highly ranked target
genes included both genes overlapping with pretrained regulon targets and
external-only targets, indicating that regulon-decoder attention recovered externally supported regulatory targets, including targets absent from the regulons used for pretraining. (Fig.~\ref{fig:regulatory_mechanism}g).

Finally, TF-target retrieval patterns varied across cellular contexts.
Representative TF programs displayed distinct enrichment patterns across
cell types, with TBX21 showing preferential retrieval in NK cells, LEF1 in
T-cell-associated populations, and HEY1 in B-lineage populations
(Fig.~\ref{fig:regulatory_mechanism}h). These context-dependent patterns
indicate that regulon supervision enables the encoder representation to retain
cell-type-associated regulatory information. Together, these analyses show that annotation and regulatory supervision
introduce complementary biological information into the shared scKITE
representation. Annotation supervision highlights genes associated with cell
identity, whereas regulon supervision supports retrieval of context-dependent
TF-associated regulatory programs.

\naturefigure
{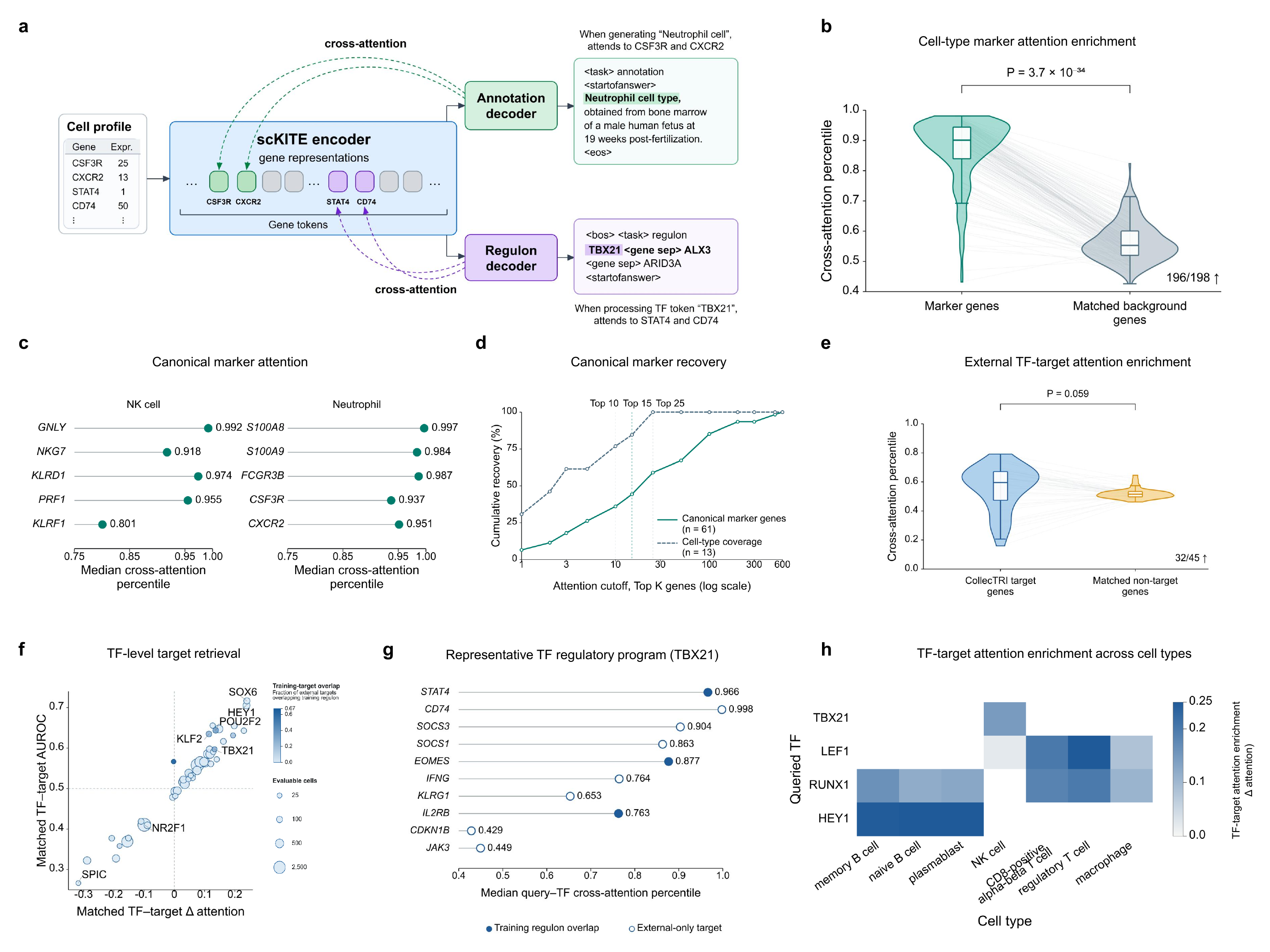}
{fig:regulatory_mechanism}
{Knowledge-guided pretraining encodes cell identity and gene regulatory information in scKITE representations.}
{
\textbf{a}, Overview of the cross-attention-based interpretation strategy for
scKITE. Decoder cross-attention between the annotation/regulon decoders and
the shared encoder representations enables identification of genes associated
with cell identity and transcription factor regulatory programs.
\textbf{b}, Enrichment of cell-type marker genes in annotation-decoder
cross-attention compared with matched background genes across evaluable cell
types.
\textbf{c}, Cross-attention ranking of canonical marker genes in representative
NK-cell and neutrophil populations.
\textbf{d}, Recovery of canonical marker genes from annotation-decoder
cross-attention rankings.
\textbf{e}, Enrichment of CollecTRI-supported TF target genes in
regulon-decoder cross-attention compared with matched non-target genes.
\textbf{f}, TF-level evaluation of regulatory target retrieval. Each point
represents an evaluated TF, with target-discrimination performance summarized
by TF-target attention enrichment and AUROC.
\textbf{g}, Representative \textit{TBX21} regulatory program revealed by
regulon-decoder cross-attention. Genes are ranked according to their
TF-associated attention, with externally supported targets highlighted.
\textbf{h}, Cell-type-dependent TF-target attention enrichment across
different cellular contexts.
}
% =========================================================
% Discussion
% =========================================================

\section{Discussion}

A central implication of our results is that knowledge-enhanced pretraining
shifts the scaling paradigm of single-cell foundation models (scFMs) beyond
data expansion alone. Increasing transcriptomic data remains valuable,
particularly when it improves biological diversity and coverage; however, our
scaling experiments reveal diminishing returns from simply expanding
pretraining data. In contrast, incorporating biological supervision
consistently improved performance across different data scales, enabling
scKITE to match or exceed representative large-scale scFMs trained on tens of
millions of samples using only around 0.18 million pretraining samples (less than
0.5\% of their data). These findings suggest that biological knowledge can
improve the data efficiency of foundation model training and provide a
complementary direction to conventional scaling strategies.

The effectiveness of knowledge-enhanced pretraining depends on matching
biological knowledge with the representations to be learned. In scKITE,
cell identity cell-associated annotations and context-specific gene regulatory programs provide complementary supervision at the cell and gene levels, respectively. This
design enables the model to capture cellular organization through cell-level
knowledge and gene regulatory context through regulatory knowledge, leading to
improved representations across cell- and gene-level tasks. More broadly, these
findings suggest that future biological foundation models should consider not
only whether biological knowledge is incorporated, but also whether the
selected knowledge is aligned with the biological properties and downstream
capabilities that the model is intended to acquire.

Several limitations should be considered when interpreting these results.
Knowledge-enhanced pretraining depends on the quality and coverage of the
biological supervision. In this study, cell identity supervision was derived
from scalable natural-language cell identity annotations associated with
single-cell profiles; however, such descriptions may vary in completeness,
specificity and consistency across datasets. Future studies could incorporate
richer cell identity knowledge from curated biological literature, expert
annotations, and other structured resources to provide more comprehensive
supervision. Similarly, the regulatory programs inferred by pySCENIC represent
predicted TF--target relationships rather than experimentally established
causal regulatory networks. The observed data-efficiency pattern was also
obtained within a specific combination of pretraining corpus, model
architecture and downstream benchmarks and should therefore not be interpreted
as a universal scaling threshold for single-cell foundation models.

Beyond the knowledge sources explored in scKITE, future biological foundation
models may benefit from integrating additional biological evidence, including
perturbation-derived causal relationships, chromatin accessibility,
TF-binding measurements, curated regulatory resources, and more standardized
experimentally grounded descriptions of cellular states. Ultimately, the
development of biologically informed foundation models will require systematic
exploration of both which biological knowledge should be incorporated and how
such knowledge should shape representation learning.

% =========================================================
% Declaration of Interests
% =========================================================
\section{Acknowledgements}

This work was supported by the National Natural Science Foundation of China (Grant No. 32673723), the China Agriculture Research System (CARS-35), and the 2115 Talent Development Program of China Agricultural University.

\section{Author Contributions}
Hanqing Zhang led the architecture design and overall study. Jie Bao contributed to model development, computational experiments, downstream task design and algorithmic evaluation. Mei Ma, Shuai Liu, Jiaying Ma, Jiaguan Liu, Jiaxiao Li, Zhenbo Li and Wenwen Gong contributed to data analysis, interpretation and manuscript revision. Zhijun Cao supervised the research and contributed to the overall study design.

\section{Declaration of Interests}

The authors declare no competing interests.

% =========================================================
% References
% =========================================================

\clearpage
\printbibliography[title={References}]

@article{aibarSCENICSinglecellRegulatory2017,
  author  = {Aibar, Sara and Gonz{\'a}lez-Blas, Carmen Bravo and Moerman, Thomas and Huynh-Thu, V{\^a}n Anh and Imrichova, Hana and Hulselmans, Gert and Rambow, Florian and Marine, Jean-Christophe and Geurts, Pierre and Aerts, Jan and van den Oord, Joost and Atak, Zeynep Kalender and Wouters, Jasper and Aerts, Stein},
  title   = {SCENIC: Single-Cell Regulatory Network Inference and Clustering},
  journal = {Nature Methods},
  year    = {2017},
  volume  = {14},
  number  = {11},
  pages   = {1083--1086},
  doi     = {10.1038/nmeth.4463}
}

@article{vandesandeScalableSCENICWorkflow2020,
  author  = {Van de Sande, Bram and Flerin, Christopher and Davie, Kristofer and De Waegeneer, Maxime and Hulselmans, Gert and Aibar, Sara and Seurinck, Ruth and Saelens, Wouter and Cannoodt, Robrecht and Rouchon, Quentin and Verbeiren, Toni and De Maeyer, Dries and Reumers, Joke and Saeys, Yvan and Aerts, Stein},
  title   = {A Scalable SCENIC Workflow for Single-Cell Gene Regulatory Network Analysis},
  journal = {Nature Protocols},
  year    = {2020},
  volume  = {15},
  number  = {7},
  pages   = {2247--2276},
  doi     = {10.1038/s41596-020-0336-2}
}

@article{theodorisTransferLearningEnables2023,
  author  = {Theodoris, Christina V. and Xiao, Ling and Chopra, Anant and Chaffin, Mark D. and Al Sayed, Zeina R. and Hill, Matthew C. and Mantineo, Helene and Brydon, Elizabeth M. and Zeng, Zexian and Liu, X. Shirley and Ellinor, Patrick T.},
  title   = {Transfer Learning Enables Predictions in Network Biology},
  journal = {Nature},
  year    = {2023},
  volume  = {618},
  number  = {7965},
  pages   = {616--624},
  doi     = {10.1038/s41586-023-06139-9}
}

@article{Cui2024scGPT,
  author  = {Cui, Haotian and Wang, Chloe and Maan, Hassaan and Pang, Kuan and Luo, Fengning and Duan, Nan and Wang, Bo},
  title   = {scGPT: Toward Building a Foundation Model for Single-Cell Multi-Omics Using Generative AI},
  journal = {Nature Methods},
  year    = {2024},
  volume  = {21},
  pages   = {1470--1480},
  doi     = {10.1038/s41592-024-02201-0}
}

@article{haoLargescaleFoundationModel2024,
  author  = {Hao, Minsheng and Gong, Jing and Zeng, Xin and Liu, Chiming and Guo, Yucheng and Cheng, Xingyi and Wang, Taifeng and Ma, Jianzhu and Zhang, Xuegong and Song, Le},
  title   = {Large-Scale Foundation Model on Single-Cell Transcriptomics},
  journal = {Nature Methods},
  year    = {2024},
  volume  = {21},
  number  = {8},
  pages   = {1481--1491},
  doi     = {10.1038/s41592-024-02305-7}
}

@article{roohaniPredictingTranscriptionalOutcomes2024,
  author  = {Roohani, Yusuf and Huang, Kexin and Leskovec, Jure},
  title   = {Predicting Transcriptional Outcomes of Novel Multigene Perturbations with GEARS},
  journal = {Nature Biotechnology},
  year    = {2024},
  volume  = {42},
  number  = {6},
  pages   = {927--935},
  doi     = {10.1038/s41587-023-01905-6}
}

@article{Schaefer2025CellWhisperer,
  author  = {Schaefer, Moritz and Peneder, Peter and Malzl, Daniel and Lombardo, Salvo Danilo and Peycheva, Mihaela and Burton, Jake and Hakobyan, Anna and Sharma, Varun and Krausgruber, Thomas and Sin, Celine and Menche, J{\"o}rg and Tomazou, Eleni M. and Bock, Christoph},
  title   = {Multimodal Learning Enables Chat-Based Exploration of Single-Cell Data},
  journal = {Nature Biotechnology},
  year    = {2025},
  doi     = {10.1038/s41587-025-02857-9}
}

@article{DenAdel2026PretrainingScale,
  author  = {DenAdel, Alan and Hughes, Madeline and Thoutam, Akshaya and Gupta, Anay and Navia, Andrew W. and Fusi, Nicolo and Raghavan, Srivatsan and Winter, Peter S. and Amini, Ava P. and Crawford, Lorin},
  title   = {Evaluating the Role of Pretraining Dataset Size and Diversity on Single-Cell Foundation Model Performance},
  journal = {Nature Methods},
  year    = {2026},
  doi     = {10.1038/s41592-026-03120-y}
}

@article{changTracingRiseBiomedical2026,
  author  = {Chang, Yuzhou and Cheng, Hao and Modi, Mirage and Wang, Guangyu and Xu, Dong and Ma, Qin},
  title   = {Tracing the Rise of Biomedical Foundation Models},
  journal = {Nature Biotechnology},
  year    = {2026},
  pages   = {1--4},
  doi     = {10.1038/s41587-026-03135-y}
}

@article{TabulaSapiens2022,
  author  = {{Tabula Sapiens Consortium}},
  title   = {The Tabula Sapiens: A Multiple-Organ, Single-Cell Transcriptomic Atlas of Humans},
  journal = {Science},
  year    = {2022},
  volume  = {376},
  number  = {6594},
  pages   = {eabl4896},
  doi     = {10.1126/science.abl4896}
}

@article{Luecken2022,
  author  = {Luecken, Malte D. and B{\"u}ttner, Maren and Chaichoompu, Kridsadakorn and Danese, Anna and Interlandi, Marta and Mueller, Michaela F. and Strobl, Daniel C. and Zappia, Luke and Dugas, Martin and Colom{\'e}-Tatch{\'e}, Maria and Theis, Fabian J.},
  title   = {Benchmarking Atlas-Level Data Integration in Single-Cell Genomics},
  journal = {Nature Methods},
  year    = {2022},
  volume  = {19},
  number  = {1},
  pages   = {41--50},
  doi     = {10.1038/s41592-021-01336-8}
}

@article{Schirmer2019,
  author  = {Schirmer, Lucas and Velmeshev, Dmitry and Holmqvist, Staffan and Kaufmann, Max and Werneburg, Sebastian and Jung, Diane and Vistnes, Stephanie and Stockley, John H. and Young, Adam and Steindel, Maike and Tung, Brian and Goyal, Nitasha and Bhaduri, Aparna and Mayer, Simone and Engler, Jan Broder and Bayraktar, Omer A. and Franklin, Robin J. M. and Haeussler, Maximilian and Reynolds, Richard and Schafer, Dorothy P. and Friese, Manuel A. and Shiow, Lawrence R. and Kriegstein, Arnold R. and Rowitch, David H.},
  title   = {Neuronal Vulnerability and Multilineage Diversity in Multiple Sclerosis},
  journal = {Nature},
  year    = {2019},
  volume  = {573},
  number  = {7772},
  pages   = {75--82},
  doi     = {10.1038/s41586-019-1404-z}
}

@article{Cheng2021,
  author  = {Cheng, Sijin and Li, Ziyi and Gao, Ranran and Xing, Baocai and Gao, Yunong and Yang, Yu and Qin, Shishang and Zhang, Lei and Ouyang, Hanqiang and Du, Peng and Jiang, Liang and Zhang, Bin and Yang, Yue and Wang, Xiliang and Ren, Xianwen and Bei, Jin-Xin and Hu, Xueda and Bu, Zhaohui and Ji, Jiafu and Zhang, Zemin},
  title   = {A Pan-Cancer Single-Cell Transcriptional Atlas of Tumor Infiltrating Myeloid Cells},
  journal = {Cell},
  year    = {2021},
  volume  = {184},
  number  = {3},
  pages   = {792--809.e23},
  doi     = {10.1016/j.cell.2021.01.010}
}

@article{DominguezConde2022,
  author  = {Dom{\'i}nguez Conde, Cecilia and Xu, Chenqu and Jarvis, Lauren B. and Rainbow, David B. and Wells, Sarah B. and Gomes, Tom{\'a}s and Howlett, Sarah K. and Suchanek, Ondrej and Polanski, Krzysztof and King, Helena W. and Mamanova, Lira and Huang, Ni and Szabo, Peter A. and Richardson, Laura and Bolt, Liam and Fasouli, Eleni S. and Mahbubani, Krishna T. and Prete, Martin and Tuck, Liz and others},
  title   = {Cross-Tissue Immune Cell Analysis Reveals Tissue-Specific Features in Humans},
  journal = {Science},
  year    = {2022},
  volume  = {376},
  number  = {6594},
  pages   = {eabl5197},
  doi     = {10.1126/science.abl5197}
}

@article{Siletti2023,
  author  = {Siletti, Kimberly and Hodge, Rebecca and Mossi Albiach, Alejandro and Lee, Kwanghun W. and Ding, Song-Lin and Hu, Lihua and L{\"o}nnerberg, Peter and Bakken, Trygve and others},
  title   = {Transcriptomic Diversity of Cell Types Across the Adult Human Brain},
  journal = {Science},
  year    = {2023},
  volume  = {382},
  number  = {6667},
  pages   = {eadd7046},
  doi     = {10.1126/science.add7046}
}

@article{CELLxGENE2025,
  author  = {{CZI Cell Science Program} and Abdulla, Shibla and Aevermann, Brian and Assis, Paulo and Badajoz, Sergio and others},
  title   = {{CZ CELLxGENE Discover}: A Single-Cell Data Platform for Scalable Exploration, Analysis and Modeling of Aggregated Data},
  journal = {Nucleic Acids Research},
  year    = {2025},
  volume  = {53},
  number  = {D1},
  pages   = {D886--D900},
  doi     = {10.1093/nar/gkae1142}
}

@article{AceraMateos2026,
  author  = {Acera-Mateos, Mario and Adiconis, Xian and Li, Jessica-Kanglin and Marchese, Domenica and Carat{\`u}, Ginevra and Hon, Chung-Chau and Tiwari, Prabha and Kojima, Miki and Vieth, Beate and others},
  title   = {Systematic Evaluation of Single-Cell Multimodal Data Integration Enhances Cell Type Resolution and Discovery of Clinically Relevant States in Complex Tissues},
  journal = {Genome Biology},
  year    = {2026},
  volume  = {27},
  pages   = {64},
  doi     = {10.1186/s13059-026-04002-4}
}

@article{Edgar2025,
  author  = {Edgar, Rachel D. and Nakib, Diana and Camat, Damra and Chung, Sai and Lumanto, Patricia and Atif, Jawairia and Perciani, Catia T. and Ma, Xue-Zhong and Thoeni, Cornelia and Selvakumaran, Nilosa and Manuel, Justin and Sayed, Blayne and Huysentruyt, Koen and Ricciuto, Amanda and McGilvray, Ian and Avitzur, Yaron and Bader, Gary D. and MacParland, Sonya A.},
  title   = {Single-Cell Atlas of Human Pediatric Liver Reveals Age-Related Hepatic Gene Signatures},
  journal = {Hepatology Communications},
  year    = {2025},
  volume  = {9},
  number  = {11},
  pages   = {e0813},
  doi     = {10.1097/HC9.0000000000000813}
}

@article{Lotfollahi2022,
  author  = {Lotfollahi, Mohammad and Naghipourfar, Mohsen and Luecken, Malte D. and Khajavi, Matin and B{\"u}ttner, Maren and Wagenstetter, Marco and Avsec, {\v Z}iga and Gayoso, Adam and Yosef, Nir and Interlandi, Marta and Rybakov, Sergei and Misharin, Alexander V. and Theis, Fabian J.},
  title   = {Mapping Single-Cell Data to Reference Atlases by Transfer Learning},
  journal = {Nature Biotechnology},
  year    = {2022},
  volume  = {40},
  number  = {1},
  pages   = {121--130},
  doi     = {10.1038/s41587-021-01001-7}
}

@article{Norman2019,
  author  = {Norman, Thomas M. and Horlbeck, Max A. and Replogle, Joseph M. and Ge, Alex Y. and Xu, Albert and Jost, Marco and Gilbert, Luke A. and Weissman, Jonathan S.},
  title   = {Exploring Genetic Interaction Manifolds Constructed from Rich Single-Cell Phenotypes},
  journal = {Science},
  year    = {2019},
  volume  = {365},
  number  = {6455},
  pages   = {786--793},
  doi     = {10.1126/science.aax4438}
}

@article{Chen2026GeneformerScaling,
  author  = {Chen, Han and Venkatesh, Madhavan S. and G{\'o}mez Ortega, Javier
             and Mahesh, Siddharth V. and Nandi, Tarak N. and Madduri, Ravi K.
             and Pelka, Karin and Theodoris, Christina V.},
  title   = {Scaling and Quantization of Large-Scale Foundation Model Enables
             Resource-Efficient Predictions in Network Biology},
  journal = {Nature Computational Science},
  year    = {2026},
  volume  = {6},
  pages   = {450--463},
  doi     = {10.1038/s43588-026-00972-4}
}

@article{dupire2026fifteen,
  author = {Dupire, Leo and Khan, Aly A. and Karaletsos, Theofanis and Kelley, Shana and Lundberg, Emma and Ma, Jian and Paull, Evan and Quake, Stephen R. and Rabadan, Raul and Rowan, Cassius and Sims, Peter and Tavazoie, Sohail and Tsang, John S. and Zhang, Mingxuan and Califano, Andrea},
  title = {Fifteen challenges for generative AI applications to cell biology},
  journal = {Cell},
  year = {2026},
  doi = {10.1016/j.cell.2026.07.004}
}

@article{Benjamini1995FDR,
  author  = {Benjamini, Yoav and Hochberg, Yosef},
  title   = {Controlling the False Discovery Rate: A Practical and Powerful Approach to Multiple Testing},
  journal = {Journal of the Royal Statistical Society: Series B (Methodological)},
  year    = {1995},
  volume  = {57},
  number  = {1},
  pages   = {289--300},
  doi     = {10.1111/j.2517-6161.1995.tb02031.x}
}

@article{Hu2023CellMarker2,
  author  = {Hu, Congxue and Li, Tengyue and Xu, Yingqi and Zhang, Xinxin and Li, Feng and Bai, Jing and Chen, Jing and Jiang, Wenqi and Yang, Kaiyue and Ou, Qi and Li, Xia and Wang, Peng and Zhang, Yunpeng},
  title   = {{CellMarker 2.0}: an updated database of manually curated cell markers in human/mouse and web tools based on {scRNA-seq} data},
  journal = {Nucleic Acids Research},
  year    = {2023},
  volume  = {51},
  number  = {D1},
  pages   = {D870--D876},
  doi     = {10.1093/nar/gkac947}
}

@article{Macosko2015DropSeq,
  title   = {Highly Parallel Genome-wide Expression Profiling of Individual Cells Using Nanoliter Droplets},
  author  = {Macosko, Evan Z. and Basu, Anindita and Satija, Rahul and Nemesh, James and Shekhar, Karthik and Goldman, Melissa and Tirosh, Itay and Bialas, Allison R. and Kamitaki, Nolan and Martersteck, Emily M. and Trombetta, John J. and Weitz, David A. and Sanes, Joshua R. and Shalek, Alex K. and Regev, Aviv and McCarroll, Steven A.},
  journal = {Cell},
  year    = {2015},
  volume  = {161},
  number  = {5},
  pages   = {1202--1214},
  doi     = {10.1016/j.cell.2015.05.002}
}

@article{Zheng2017MassivelyParallel,
  title   = {Massively parallel digital transcriptional profiling of single cells},
  author  = {Zheng, Grace X. Y. and Terry, Jessica M. and Belgrader, Phillip and others},
  journal = {Nature Communications},
  year    = {2017},
  volume  = {8},
  pages   = {14049},
  doi     = {10.1038/ncomms14049}
}

@article{Regev2017HumanCellAtlas,
  title   = {The Human Cell Atlas},
  author  = {Regev, Aviv and Teichmann, Sarah A. and Lander, Eric S. and others},
  journal = {eLife},
  year    = {2017},
  volume  = {6},
  pages   = {e27041},
  doi     = {10.7554/eLife.27041}
}

@article{Yang2022scBERT,
  title   = {scBERT as a large-scale pretrained deep language model for cell type annotation of single-cell RNA-seq data},
  author  = {Yang, Fan and Wang, Wenchuan and Wang, Fang and Fang, Yuan and Tang, Duyu and Huang, Junzhou and Lu, Hui and Yao, Jianhua},
  journal = {Nature Machine Intelligence},
  year    = {2022},
  volume  = {4},
  pages   = {852--866},
  doi     = {10.1038/s42256-022-00534-z}
}

@inproceedings{Wen2024CellPLM,
  title     = {CellPLM: Pre-Training of Cell Language Model Beyond Single Cells},
  author    = {Wen, Hongzhi and Tang, Wenzhuo and Dai, Xinnan and Ding, Jiayuan and Jin, Wei and Xie, Yuying and Tang, Jiliang},
  booktitle = {International Conference on Learning Representations},
  year      = {2024}
}

@article{Heimberg2025SCimilarity,
  title   = {A cell atlas foundation model for scalable search of similar human cells},
  author  = {Heimberg, Graham and Kuo, Tony and DePianto, Daryle J. and others},
  journal = {Nature},
  year    = {2025},
  volume  = {638},
  number  = {8052},
  pages   = {1085--1094},
  doi     = {10.1038/s41586-024-08411-y}
}

@article{Rosen2026UCE,
  title   = {Universal cell embedding provides a foundation model for cell biology},
  author  = {Rosen, Yanay and Roohani, Yusuf and Agrawal, Ayush and Samotor{\v{c}}an, Leon and {Tabula Sapiens Consortium} and Quake, Stephen R. and Leskovec, Jure},
  journal = {Nature},
  year    = {2026},
  volume  = {656},
  number  = {8126},
  pages   = {183--191},
  doi     = {10.1038/s41586-026-10689-z}
}

@article{Gandhi2025TahoeX1,
  title   = {Tahoe-x1: Scaling Perturbation-Trained Single-Cell Foundation Models to 3 Billion Parameters},
  author  = {Gandhi, Shreshth and Javadi, Farnoosh and Svensson, Valentine and Khan, Umair and Jones, Matthew G. and Yu, John and Merico, Daniele and Goodarzi, Hani and Alidoust, Nima},
  journal = {bioRxiv},
  year    = {2025},
  doi     = {10.1101/2025.10.23.683759}
}

@misc{Kaplan2020ScalingLaws,
  title         = {Scaling Laws for Neural Language Models},
  author        = {Kaplan, Jared and McCandlish, Sam and Henighan, Tom and Brown, Tom B. and Chess, Benjamin and Child, Rewon and Gray, Scott and Radford, Alec and Wu, Jeffrey and Amodei, Dario},
  year          = {2020},
  eprint        = {2001.08361},
  archivePrefix = {arXiv},
  primaryClass  = {cs.LG}
}

@inproceedings{Zhao2024LangCell,
  title     = {LangCell: Language-Cell Pre-training for Cell Identity Understanding},
  author    = {Zhao, Suyuan and Zhang, Jiahuan and Wu, Yushuai and Luo, Yizhen and Nie, Zaiqing},
  booktitle = {Proceedings of the 41st International Conference on Machine Learning},
  series    = {Proceedings of Machine Learning Research},
  volume    = {235},
  pages     = {61159--61185},
  year      = {2024},
  publisher = {PMLR}
}

@inproceedings{Levine2024Cell2Sentence,
  title     = {Cell2Sentence: Teaching Large Language Models the Language of Biology},
  author    = {Levine, Daniel and Rizvi, Syed A. and L{\'e}vy, Sacha and Pallikkavaliyaveetil, Nazreen and Zhang, David and Chen, Xingyu and Ghadermarzi, Sina and Wu, Ruiming and Zheng, Zihe and Vrkic, Ivan and Zhong, Anna and Raskin, Daphne and Han, Insu and De Oliveira Fonseca, Antonio Henrique and Ortega Caro, Josue and Karbasi, Amin and Dhodapkar, Rahul Madhav and Van Dijk, David},
  booktitle = {Proceedings of the 41st International Conference on Machine Learning},
  series    = {Proceedings of Machine Learning Research},
  volume    = {235},
  pages     = {27299--27325},
  year      = {2024},
  publisher = {PMLR}
}

@article{BravoGonzalezBlas2023SCENICPlus,
  title   = {SCENIC+: single-cell multiomic inference of enhancers and gene regulatory networks},
  author  = {Bravo Gonz{\'a}lez-Blas, Carmen and De Winter, Seppe and Hulselmans, Gert and Hecker, Nikolai and Matetovici, Irina and Christiaens, Valerie and Poovathingal, Suresh and Wouters, Jasper and Aibar, Sara and Aerts, Stein},
  journal = {Nature Methods},
  year    = {2023},
  volume  = {20},
  number  = {9},
  pages   = {1355--1367},
  doi     = {10.1038/s41592-023-01938-4}
}

@article{Moerman2019GRNBoost2,
  title   = {GRNBoost2 and Arboreto: efficient and scalable inference of gene regulatory networks},
  author  = {Moerman, Thomas and Aibar Santos, Sara and Bravo Gonz{\'a}lez-Blas, Carmen and Simm, Jaak and Moreau, Yves and Aerts, Jan and Aerts, Stein},
  journal = {Bioinformatics},
  year    = {2019},
  volume  = {35},
  number  = {12},
  pages   = {2159--2161},
  doi     = {10.1093/bioinformatics/bty916}
}

@article{MullerDott2023CollecTRI,
  title   = {Expanding the coverage of regulons from high-confidence prior knowledge for accurate estimation of transcription factor activities},
  author  = {M{\"u}ller-Dott, Sophia and Tsirvouli, Eirini and Vazquez, Miguel and Ramirez Flores, Ricardo O. and Badia-I-Mompel, Pau and Fallegger, Robin and T{\"u}rei, D{\'e}nes and L{\ae}greid, Astrid and Saez-Rodriguez, Julio},
  journal = {Nucleic Acids Research},
  year    = {2023},
  volume  = {51},
  number  = {20},
  pages   = {10934--10949},
  doi     = {10.1093/nar/gkad841}
}

@article{Hie2019GeometricSketching,
  title   = {Geometric Sketching Compactly Summarizes the Single-Cell Transcriptomic Landscape},
  author  = {Hie, Brian and Cho, Hyunghoon and DeMeo, Benjamin and Bryson, Bryan and Berger, Bonnie},
  journal = {Cell Systems},
  year    = {2019},
  volume  = {8},
  number  = {6},
  pages   = {483--493.e7},
  doi     = {10.1016/j.cels.2019.05.003}
}

@inproceedings{Loshchilov2019AdamW,
  title     = {Decoupled Weight Decay Regularization},
  author    = {Loshchilov, Ilya and Hutter, Frank},
  booktitle = {International Conference on Learning Representations},
  year      = {2019}
}

@article{Traag2019Leiden,
  title   = {From Louvain to Leiden: guaranteeing well-connected communities},
  author  = {Traag, V. A. and Waltman, L. and van Eck, N. J.},
  journal = {Scientific Reports},
  year    = {2019},
  volume  = {9},
  pages   = {5233},
  doi     = {10.1038/s41598-019-41695-z}
}

@article{McInnes2018UMAP,
  title   = {UMAP: Uniform Manifold Approximation and Projection},
  author  = {McInnes, Leland and Healy, John and Saul, Nathaniel and Gro{\ss}berger, Lukas},
  journal = {Journal of Open Source Software},
  year    = {2018},
  volume  = {3},
  number  = {29},
  pages   = {861},
  doi     = {10.21105/joss.00861}
}

@article{stuart2019comprehensive,
  title={Comprehensive integration of single-cell data},
  author={Stuart, Tim and Butler, Andrew and Hoffman, Paul and Hafemeister, Christoph and Papalexi, Efthymia and Mauck III, William M and Hao, Yuhan and Stoeckius, Marlon and Smibert, Peter and Satija, Rahul},
  journal={Cell},
  volume={177},
  number={7},
  pages={1888--1902},
  year={2019},
  publisher={Elsevier}
}

@article{wilcoxon1945individualcomparisons,
  title={Individual comparisons by ranking methods},
  author={Wilcoxon, Frank},
  journal={Biometrics Bulletin},
  volume={1},
  number={6},
  pages={80--83},
  year={1945}
}

@article{Replogle2022,
  title={A scalable platform for the development of single-cell perturbation screens},
  author={Replogle, Joseph M. and Saunders, Robert A. and Pogson, Anna N. and Hussmann, Jacob A. and Lenail, Adam and Gopalan, Harish and Li, Ying and Strittmatter, Kai and Dele-Ojo, Bamidele and Naik, Pooja and others},
  journal={Nature Genetics},
  volume={54},
  pages={1418--1429},
  year={2022},
  publisher={Springer Nature},
  doi={10.1038/s41588-022-01103-7}
}

@article{Wei2026Benchmark,
  title={Benchmarking algorithms for generalizable single-cell perturbation response prediction},
  author={Wei, Zhiting and Wang, Yiheng and Gao, Yicheng and Wang, Shuguang and Li, Ping and Si, Duanmiao and Gao, Yuli and Wu, Siqi and Li, Danlu and Dong, Kejing and Yang, Xingbo and Tang, Chen and Fu, Shaliu and Chen, Xiaohan and Li, Wannian and You, Yuzhou and Zhang, Chen and Liang, Aibin and Chuai, Guohui and Liu, Qi},
  journal={Nature Methods},
  volume={23},
  pages={451--464},
  year={2026},
  publisher={Springer Nature},
  doi={10.1038/s41592-025-02980-0}
}

@article{Chan2024CELLxGENE,
  title={The {CELLxGENE} {Census}: a large-scale reference dataset for single-cell genomics},
  author={Chan, Z. and others},
  journal={Nature Methods},
  year={2024},
  volume={21},
  pages={209--218},
  doi={10.1038/s41592-023-02167-9}
}

% =========================================================
% Supplementary Methods
% =========================================================

\clearpage

% Enable section numbering for Supplementary Methods
\setcounter{secnumdepth}{3}

% Reset counters
\setcounter{section}{0}
\setcounter{subsection}{0}
\setcounter{subsubsection}{0}

% Numbering format
\renewcommand{\thesection}{\arabic{section}}
\renewcommand{\thesubsection}{\thesection.\arabic{subsection}}
\renewcommand{\thesubsubsection}{\thesubsection.\arabic{subsubsection}}

% Supplementary Methods title
\begin{center}
    {\Large\bfseries Supplementary Methods}
\end{center}

\vspace{1em}

% =========================================================
% Methods body
% =========================================================

% =========================================================
% 1. Pretraining data construction
% =========================================================

\section{Pretraining data construction}

\subsection{Transcriptomic pretraining corpus}
\label{sec:pretraining_corpus}

The transcriptomic pretraining corpus was derived from the CELLxGENE component of the CellWhisperer dataset \cite{Schaefer2025CellWhisperer}. In the original CellWhisperer pipeline, cells within individual CELLxGENE datasets were grouped according to available metadata and averaged to generate pseudo-bulk transcriptomic profiles, yielding 376,983 human transcriptome--annotation pairs. For scKITE, we extracted the gene identities, corresponding expression values and matched natural-language annotations from these profiles. The corpus was divided into 358,134 training profiles and 18,849 profiles retained as a fixed validation set.

\subsection{Natural-language annotations}
\label{sec:annotation_supervision}
Natural-language annotations were obtained directly from the CellWhisperer-derived CELLxGENE corpus \cite{Schaefer2025CellWhisperer}; no additional large language model was used to regenerate them. In the original CellWhisperer curation pipeline, metadata associated with each pseudo-bulk profile were condensed into concise biological descriptions containing information such as cell identity, tissue or organ of origin, disease or physiological condition and available donor characteristics. These pre-generated descriptions were used as annotation supervision during Stage~2 pretraining.

\subsection{Regulatory knowledge construction}
\label{sec:regulon_supervision}
Regulatory supervision was constructed using pySCENIC following its standard
workflow of co-expression network inference, cis-regulatory motif enrichment
and regulon activity scoring
\cite{aibarSCENICSinglecellRegulatory2017,vandesandeScalableSCENICWorkflow2020}.
GRNBoost2 was first used to infer candidate transcription factor (TF)--target
associations and corresponding interaction importance scores
\cite{Moerman2019GRNBoost2}. The resulting regulatory modules were refined
using cisTarget motif enrichment. Motif-supported targets associated with the
same TF were merged to define a single TF-centered regulon. When the same
TF--target relationship was supported by multiple motif-derived sets, the
maximum GRNBoost2 importance score was retained. Targets within each regulon
were ordered by decreasing interaction importance.

Regulon activity was quantified for each transcriptomic profile using AUCell.
A regulon-specific activity threshold was estimated from the distribution of
AUCell scores across profiles. Regulon $r$ was considered active in profile $c$
when
\begin{equation}
\mathrm{AUC}_{c,r} > \tau_r,
\end{equation}
where $\tau_r$ denotes the regulon-specific activity threshold. Regulons with
invariant AUCell scores across profiles were excluded, leaving 530 informative
regulons. Each retained regulon was assigned a unique identifier in a global
regulon lookup table, and each transcriptomic profile retained its active
regulon identifiers and corresponding AUCell scores.

TFs and target genes were mapped to the global gene vocabulary. During
Stage~2 data construction, a profile-specific target set was generated for each
active regulon by intersecting its globally defined target set with genes
expressed in the corresponding transcriptomic profile, while preserving the
global interaction-importance ordering. Active regulons with no remaining
expressed targets were excluded from decoder supervision for that profile.

During training, up to three eligible active regulons were selected for each
profile using an epoch-aware cyclic-without-replacement strategy. For each
profile, a stable permutation of eligible regulons was defined, and the
selection window was shifted across epochs so that different regulons were
presented over training without duplication within a given selection.
Validation used a fixed deterministic selection.

The TFs of the selected regulons first formed a query sequence separated by
\texttt{<gene\_sep>} tokens. The sequence was preceded by the
\texttt{<task>} token and the task identifier \texttt{regulon}, followed by
\texttt{<startofanswer>}. The corresponding regulons were then serialized
autoregressively, with each TF followed by a \texttt{<arrow>} token and its
importance-ranked profile-specific target genes. Individual regulons were
separated by \texttt{<regulon\_sep>}, and the complete sequence terminated
with \texttt{<eos>}:

\[
\begin{aligned}
&\texttt{<task> regulon}\;
\mathrm{TF}_1\;
\texttt{<gene\_sep>}\;
\mathrm{TF}_2\;
\texttt{<gene\_sep>}\;
\mathrm{TF}_3\;
\texttt{<startofanswer>}\\
&\mathrm{TF}_1\;
\texttt{<arrow>}\;
g_{1,1},\ldots,g_{1,n_1}\;
\texttt{<regulon\_sep>}\\
&\mathrm{TF}_2\;
\texttt{<arrow>}\;
g_{2,1},\ldots,g_{2,n_2}\;
\texttt{<regulon\_sep>}\\
&\mathrm{TF}_3\;
\texttt{<arrow>}\;
g_{3,1},\ldots,g_{3,n_3}\;
\texttt{<eos>}.
\end{aligned}
\]

When the complete target sequence exceeded the available decoder context,
the target-token budget was distributed across the selected regulons in a
round-robin manner while preserving the within-regulon interaction-importance
ordering.

% =========================================================
% 2. Downstream benchmark datasets
% =========================================================

\section{Downstream benchmark datasets}

\subsection{Cell type classification datasets}

\paragraph{Tabula Sapiens.}
The Tabula Sapiens dataset was obtained from the publicly available Tabula Sapiens human cell atlas \cite{TabulaSapiens2022}. Cells from blood, bone marrow, lung, mammary and thymus with valid cell type annotations were retained, yielding 143,133 cells across 75 cell types. Cells were stratified by cell type and divided into training, validation and test sets at a ratio of 80\%, 10\% and 10\%, respectively.

\paragraph{Human pancreas.}
The human pancreas dataset was obtained from a publicly available processed human pancreas single-cell dataset \cite{Luecken2022}, comprising five independent scRNA-seq studies and 16,382 cells across 14 cell types. Cells from two studies were used to construct the training and validation sets, containing 11,705 and 2,347 cells, respectively, whereas cells from the remaining three studies formed the 2,330-cell test set.

\paragraph{Multiple sclerosis.}
The multiple sclerosis dataset was originally obtained from the EMBL-EBI Single Cell Expression Atlas under accession E-HCAD-35 \cite{Schirmer2019}; we used the preprocessed version released with scGPT \cite{Cui2024scGPT}. After removal of cell types present only in the MS subset, the dataset contained 21,312 cells across 18 cell types, including 7,844 cells from nine healthy controls and 13,468 cells from 12 individuals with MS. Seven healthy individuals were used for training, two for validation and all 12 MS individuals for testing.

\paragraph{Tumor-infiltrating myeloid.}
The tumor-infiltrating myeloid dataset was originally obtained from GEO accession GSE154763 \cite{Cheng2021}; we used the preprocessed version released with scGPT \cite{Cui2024scGPT}. Six cancer-associated groups (UCEC, PAAD, THCA, LYM, cDC2 and kidney) formed the reference subset and three groups (MYE, OV-FTC and ESCA) formed the query subset. The reference subset contained 9,748 cells and was divided 80:20 into training and validation sets; all 3,430 query cells were retained for testing.

\paragraph{Cross-tissue Immune Cell Atlas.}
The cross-tissue immune dataset was obtained from the Cross-tissue Immune Cell Atlas \cite{DominguezConde2022}. The processed dataset contained 324,458 cells from nine tissue environments and 43 annotated immune-cell types. Six tissues (252,520 cells) were used for training, lung (35,419 cells) for validation, and liver together with mesenteric lymph node (36,519 cells) for testing.

\subsection{Batch integration datasets}

\paragraph{Perirhinal cortex.}
The perirhinal cortex dataset was derived from the adult human brain transcriptomic atlas of Siletti et al. \cite{Siletti2023}; we used the processed version released with scGPT \cite{Cui2024scGPT}. Two assay-defined technical batches, \texttt{10X222\_1} and \texttt{10X222\_2}, contained 8,465 and 9,070 cells, respectively. The dataset contained 17,535 cells, 59,357 genes and 10 annotated cell types and was used for cross-assay integration.

\paragraph{Renal.}
The renal dataset was obtained from CZ CELLxGENE Discover \cite{CELLxGENE2025} for the multimodal renal cortex benchmark \cite{AceraMateos2026}. It contained 97,125 cells from 19 donors across 35 annotated cell types. Cells were grouped by assay into 10x 3$'$ v3, 10x 5$'$ v1 and 10x multiome batches containing 35,513, 7,813 and 53,799 cells, respectively, and were used for cross-assay integration.

\paragraph{Liver.}
The liver dataset was obtained from CZ CELLxGENE Discover \cite{CELLxGENE2025} and corresponds to the human pediatric and adult liver atlas of Edgar et al. \cite{Edgar2025}. The dataset contained 69,032 cells from 16 donors, including nine pediatric and seven adult donors, across 19 annotated cell types. Donor identity was used as the batch variable for cross-donor integration. Because liver-lobe distribution was imbalanced across donors, results were interpreted as reduction of donor-associated variation while preserving cell type structure rather than complete removal of all donor-related biological differences.

\paragraph{COVID-19.}
The COVID-19 dataset was derived from the single-cell reference-mapping dataset assembled by Lotfollahi et al. \cite{Lotfollahi2022}, and we used the processed version released with scGPT \cite{Cui2024scGPT}. The original integrated dataset contained 274,346 cells and 18,474 genes from lung tissue, PBMCs and bone marrow and was subsampled to 20,000 cells. Following the scGPT benchmark, the processed dataset was organized into 18 distinct study/sample-level batches, which were used as batch identities for integration evaluation.

\subsection{Perturbation prediction datasets}

\paragraph{Norman.}
The Norman dataset was obtained from the CRISPRa Perturb-seq study of Norman et al. \cite{Norman2019}. It contains K562 cells subjected to 105 single-gene and 131 two-gene perturbations, with each perturbation measured in approximately 300--700 cells. Perturbation conditions were partitioned according to the GEARS simulation split for evaluation of seen and unseen perturbations.

\paragraph{Replogle.}
The Replogle dataset was derived from the single-cell CRISPR-interference perturbation experiments of Replogle et al. \cite{Replogle2022}. We used the processed version included in the perturbation-generalization benchmark of Wei et al. \cite{Wei2026Benchmark}, which was obtained from the PerturBase database. Replogle represents one of the single-gene perturbation experiments from the original study and contains 105 genetic perturbations and 102,148 cells after benchmark preprocessing. This dataset was used to evaluate model generalization to previously unseen single-gene perturbations.

% =========================================================

% =========================================================
% 3. Model architecture and representations
% =========================================================

\section{Model architecture and representations}

\subsection{Model input}

scKITE used a unified global vocabulary containing gene tokens, text tokens and
task-specific structural tokens. Gene symbols or Ensembl identifiers were mapped
to global gene-token IDs, while natural-language annotations were tokenized with
the biomedical text tokenizer and mapped into the same global ID space.
Structural tokens included \texttt{<cls>}, \texttt{<task>},
\texttt{<startofanswer>}, \texttt{<arrow>}, \texttt{<gene\_sep>},
\texttt{<regulon\_sep>} and \texttt{<eos>}.

For each transcriptomic profile, genes with expression values greater than zero
were retained and ordered in descending expression order, with ties resolved by
ascending gene-token ID. Up to 2,048 expressed genes were retained and a
\texttt{<cls>} token was prepended, yielding a maximum sequence length of 2,049
tokens. Positive expression values were quantile-binned independently within
each profile into 50 non-zero levels, with zero reserved for non-expressed
values, resulting in 51 expression levels in total.

For masked-expression reconstruction, gene identities were retained at selected
positions while their expression values were replaced by a dedicated mask value.
The \texttt{<cls>} and padding positions were excluded from masking. The masking
probability was 30\% in Stage~1 and 10\% in Stage~2.

For each input position $i$, gene identity, expression level and masking status
were combined by element-wise addition,
\begin{equation}
\mathbf{x}_i =
\mathbf{e}^{\mathrm{gene}}_i +
\mathbf{e}^{\mathrm{expr}}_i +
\mathbf{e}^{\mathrm{mask}}_i,
\end{equation}
where gene identities were represented by learnable gene embeddings, expression
values were projected into the model hidden space by a multilayer value encoder,
and a learnable binary mask embedding indicated whether the corresponding
expression value had been masked.

\subsection{Encoder}
\label{sec:encoder}
The scKITE backbone consists of 12 pre-normalization Transformer encoder blocks with a hidden dimension of 512 and eight attention heads. Each block contains multi-head self-attention followed by a position-wise feed-forward network with an intermediate dimension four times the hidden size. Residual connections are applied around both modules, with GELU activation and dropout of 0.1. A final layer-normalization operation is applied after the last block. For an input sequence $\mathbf{X}$, the encoder produces contextualized hidden states
\begin{equation}
\mathbf{H}=
[\mathbf{h}_{\mathrm{cls}},\mathbf{h}_1,\ldots,\mathbf{h}_n]
=\mathrm{Encoder}(\mathbf{X}).
\end{equation}

\subsection{Knowledge decoders}
\label{sec:decoders}
Stage~2 introduces two task-specific autoregressive decoders: a regulon decoder and an annotation decoder. The decoders have the same architecture but independent parameters. Each consists of two Transformer decoder blocks with a hidden dimension of 512 and eight attention heads. Each block contains causal self-attention, encoder--decoder cross-attention and a feed-forward network. Decoder hidden states act as cross-attention queries, whereas the complete encoder hidden-state sequence $\mathbf{H}$ provides keys and values. The regulon decoder generates profile-specific TF--target sequences and the annotation decoder generates the paired natural-language description. The two branches use independent positional embeddings and output heads.

\subsection{Learned representations}
\label{sec:model_output}
The final encoder hidden state corresponding to \texttt{<cls>} was used as the cell representation,
\begin{equation}
\mathbf{z}_{\mathrm{cell}}=\mathbf{h}_{\mathrm{cls}},
\end{equation}
whereas the hidden state associated with each expressed gene was retained as its contextual gene representation,
\begin{equation}
\mathbf{z}_{g_i}=\mathbf{h}_i.
\end{equation}
Because token-level hidden states depend on the complete transcriptomic context, the same gene can acquire different representations across cellular states. Cell embeddings were used for cell-level analyses, whereas contextual gene embeddings were used for perturbation prediction and gene-level regulatory analyses.

% =========================================================

% =========================================================
% 4. Two-stage pretraining
% =========================================================

\section{Two-stage pretraining strategy}

\subsection{Stage 1: transcriptomic self-supervised pretraining}
\label{sec:stage1_pretraining}
Stage~1 trained the shared Transformer encoder using masked-expression reconstruction alone. For each profile, 30\% of valid gene positions were selected for masking. Gene identities were retained, whereas their binned expression values were replaced by the mask value. A regression head predicted the original expression levels from the corresponding encoder hidden states. The reconstruction loss was
\begin{equation}
\mathcal{L}_{\mathrm{expr}}
=
\frac{1}{|\mathcal{M}|}
\sum_{i\in\mathcal{M}}
(\hat{x}_i-x_i)^2,
\end{equation}
where $\mathcal{M}$ denotes the masked positions. The resulting Stage~1 encoder checkpoint was used to initialize Stage~2.

\subsection{Stage 2: knowledge-enhanced pretraining}
\label{sec:stage2_pretraining}
Stage~2 initialized the encoder from the Stage~1 checkpoint and jointly
optimized the trainable encoder, regulon decoder and annotation decoder.
Masked-expression reconstruction was retained with the masking probability
reduced to 10\%. In parallel, the regulon decoder generated profile-specific
regulon sequences and the annotation decoder generated paired natural-language
descriptions. Both decoder objectives were optimized using autoregressive token-level cross-entropy losses with teacher forcing, and were jointly optimized with the masked-expression reconstruction objective.

For the annotation decoder, the loss was defined as

\begin{equation}
\mathcal{L}_{\mathrm{ann}}
=
-\frac{1}{T_{\mathrm{ann}}}
\sum_{t=1}^{T_{\mathrm{ann}}}
\log
P(y_t^{\mathrm{ann}}
|
y_{<t}^{\mathrm{ann}},H),
\end{equation}

where $H$ denotes encoder hidden states and $y_t^{\mathrm{ann}}$ denotes the
$t$-th annotation token.

Similarly, the regulon decoder objective was defined as

\begin{equation}
\mathcal{L}_{\mathrm{reg}}
=
-\frac{1}{T_{\mathrm{reg}}}
\sum_{t=1}^{T_{\mathrm{reg}}}
\log
P(y_t^{\mathrm{reg}}
|
y_{<t}^{\mathrm{reg}},H),
\end{equation}

where $y_t^{\mathrm{reg}}$ represents the $t$-th regulon token.

The three objectives were jointly optimized using the following Stage~2
training objective:

\begin{equation}
\mathcal{L}_{\mathrm{Stage2}}
=
\lambda_{\mathrm{expr}}\mathcal{L}_{\mathrm{expr}}
+
\lambda_{\mathrm{reg}}\mathcal{L}_{\mathrm{reg}}
+
\lambda_{\mathrm{ann}}\mathcal{L}_{\mathrm{ann}},
\end{equation}

where $\lambda_{\mathrm{expr}}$, $\lambda_{\mathrm{reg}}$ and
$\lambda_{\mathrm{ann}}$ control the relative contributions of transcriptomic
reconstruction, regulon supervision and annotation supervision, respectively.
All loss weights were set to 1 in our experiments.

\subsection{Encoder extraction after pretraining}

The regulon and annotation decoders were used only during Stage~2 pretraining. After pretraining, both decoders were discarded and downstream applications used only the shared scKITE encoder. Consequently, downstream inference requires only transcriptomic input and does not require natural-language annotations or regulon targets.

% =========================================================

% =========================================================
% 5. Pretraining data scaling analysis
% =========================================================

\section{Pretraining data scaling analysis}

To assess how pretraining data scale affected downstream performance, we
constructed a series of nested subsets from the full training pool. The
training pool contained 358,134 transcriptomic profiles, while a fixed
validation set of 18,849 profiles was held out from all subset construction
and used consistently across scaling experiments. We evaluated pretraining
fractions of 10\%, 25\%, 50\% and 100\%, corresponding to 35,813, 89,534,
179,067 and 358,134 training profiles, respectively.

To reduce transcriptomic redundancy while preserving the global structure
of the expression space, subsets were constructed using nested geometric
sketching \cite{Hie2019GeometricSketching}. Raw expression profiles were
library-size normalized to 10,000 counts per profile and log-transformed.
The 4,000 most variable genes were retained, and the expression matrix was
projected into a 50-dimensional latent space using truncated singular value
decomposition (SVD). Geometric sketching was then applied in this shared
space to select representative profiles spanning the transcriptomic
manifold. A fixed hierarchy of geometric regions and within-region sample
rankings was used to generate strictly nested subsets,

\[
S_{10\%}
\subset
S_{25\%}
\subset
S_{50\%}
\subset
S_{100\%}.
\]

Thus, increasing the pretraining fraction added new profiles without
replacing profiles retained at smaller fractions.

For each data fraction, Stage~1 pretraining was performed independently from
random initialization using the corresponding subset. Stage~2
knowledge-enhanced pretraining was then initialized from the matched
Stage~1 checkpoint. Model architecture, masking strategy, optimization
settings and validation data were held constant across data fractions.
Rather than comparing models after an identical number of epochs, training
was continued until validation performance approached convergence, reducing
the confounding effect of fewer optimization steps in smaller datasets.

Scaling behaviour was evaluated using three representative downstream
tasks: cell type annotation, batch integration and genetic
perturbation-response prediction. Cell type annotation performance was
summarized as the mean of accuracy, macro-F1, precision and recall.
Perturbation-response prediction was summarized using the overall Top-20
differentially expressed gene mean squared error (Top-20 DE MSE). Batch
integration performance was summarized using a composite integration score,

\[
P_{\mathrm{integration}}
=
0.6\,\mathrm{AvgBIO}
+
0.4\,\mathrm{AvgBATCH},
\]

which assigns greater weight to biological conservation while retaining
batch-mixing performance.

To compare scaling trends across tasks with different metric ranges and
directions, each task-specific score was normalized to the corresponding
full-data scKITE result. For metrics in which larger values indicate better
performance,

\[
P_{\mathrm{norm}}(f)
=
\frac{P(f)}
{P_{\mathrm{scKITE},100\%}},
\]

whereas for error metrics in which smaller values indicate better
performance,

\[
P_{\mathrm{norm}}(f)
=
\frac{P_{\mathrm{scKITE},100\%}}
{P(f)}.
\]

Here, \(f\) denotes the fraction of the full pretraining corpus. The same
full-data scKITE reference was used for both scKITE and scKITE (w/o knowledge enhancement). The normalized scores from the three downstream tasks
were then averaged with equal weight to obtain the mean normalized
downstream performance used for scaling analysis.
To summarize performance across the three downstream tasks, we further
defined the mean normalized downstream performance at each pretraining
fraction as

\[
P_{\mathrm{overall}}(f)
=
\frac{1}{3}
\left[
P_{\mathrm{norm}}^{\mathrm{annotation}}(f)
+
P_{\mathrm{norm}}^{\mathrm{integration}}(f)
+
P_{\mathrm{norm}}^{\mathrm{perturbation}}(f)
\right].
\]

The three tasks were assigned equal weight. This overall score was calculated
separately for scKITE and scKITE (w/o knowledge enhancement) at each pretraining
fraction, using the full-data scKITE model as the common normalization
reference. The resulting scores were used to generate the two scaling curves
shown in Fig.~\ref{fig:framework}a.

% =========================================================

% =========================================================
% 6. Downstream evaluation
% =========================================================

\section{Downstream evaluation}

% =========================================================
% Cell type annotation and representation analysis
% =========================================================

\subsection{Cell type classification and representation analysis}
\label{sec:cell_annotation}
Cell type annotation was evaluated under zero-shot and full fine-tuning
settings using the predefined training, validation and test partitions of each
benchmark dataset. The label space was defined by the cell types present in
the training split, and validation or test cells with labels absent from the
training label space were excluded from evaluation.

For zero-shot annotation, the pretrained encoder was frozen and no trainable
classification head or downstream optimization was introduced. The
\texttt{<cls>} embedding of each training cell was used as the labeled
reference set, whereas test cells were treated as queries. Each query was
assigned the majority label among its five nearest reference embeddings using
Euclidean distance in the unnormalized embedding space. The validation split
was not used in this setting. Here, zero-shot annotation refers to prediction
without parameter updates or a trained downstream neural classifier, while
labeled training cells were used as references for the non-parametric
nearest-neighbor classifier.

For full fine-tuning, a randomly initialized two-layer multilayer perceptron
was attached to the \texttt{<cls>} representation, and the pretrained encoder
and classification head were jointly optimized using multiclass
cross-entropy. The classification head consisted of layer normalization, a
linear layer, GELU activation, dropout, a second layer-normalization operation
and a final linear classifier. Models were trained for up to 10 epochs using
AdamW, mixed precision and gradient clipping
\cite{Loshchilov2019AdamW}. The checkpoint with the lowest validation
cross-entropy loss was evaluated once on the held-out test set. Performance
under both settings was assessed using accuracy, macro-precision,
macro-recall and macro-F1. Macro-averaged metrics were calculated across cell
types represented in the test split, with undefined class-level values set to
zero.

To further characterize the biological organization captured by the
pretrained representations, cell embeddings were extracted from the final
encoder \texttt{<cls>} state without task-specific fine-tuning. These analyses
used manually curated annotations from the cross-tissue immune dataset, with
doublets excluded. scKITE (w/o knowledge enhancement) and scKITE models were evaluated using
identical cells and identical analysis procedures. Broad immune-cell
categories defined by the dataset annotations are referred to here as immune
compartments.

Three annotated immune compartments were considered: the B cell compartment,
T and innate lymphoid cells, and the myeloid compartment. Six abundant,
manually curated cell types were selected from each compartment, yielding 18
representative cell types. Cell-type centroids were calculated from the
corresponding cell embeddings, and pairwise cosine distances between centroids
were subjected to hierarchical clustering using average linkage with optimal
leaf ordering. Each dendrogram was partitioned into three clusters, and the
agreement between the resulting unsupervised clusters and the annotated immune
compartments was quantified using adjusted Rand index (ARI) and normalized
mutual information (NMI).

At the individual-cell level, cell pairs were grouped into three levels of
biological relatedness: cells of the same type, cells of different types
within the same immune compartment, and cells from different immune
compartments. Up to 80 cells were randomly sampled from each cell type, and
12,000 cell pairs were sampled for each relationship level using a fixed
random seed. Identical cell pairs were used for both models. Cell embeddings
were $\ell_2$-normalized, pairwise cosine distances were calculated, and the
resulting distributions were estimated using Gaussian kernel density
estimation and visualized as ridgeline plots.

Directional cross-tissue retrieval was further evaluated across the nine
tissues represented in the processed cross-tissue immune dataset. For each
cell type within each tissue, a centroid was calculated from the corresponding
cell embeddings. A cell-type centroid from a source tissue was used as the
query and compared with cell-type centroids in a target tissue using cosine
similarity. Candidate cell types were restricted to the same annotated immune
compartment as the query to evaluate fine-grained cell identity beyond broad
compartment separation. Comparisons were retained only when the query cell
type was represented in both tissues and at least two candidate cell types
were available in the target tissue. Retrieval was considered correct when
the most similar target centroid corresponded to the same annotated cell type
as the query. Top-1 retrieval accuracy was calculated separately for each
directional source--target tissue pair, with self-tissue comparisons excluded.

% =========================================================
% Batch integration
% =========================================================

\subsection{Batch integration}
\label{sec:batch_integration}
Batch integration was evaluated to determine whether pretrained
representations preserved biological cell type structure while reducing
batch-associated variation. The four benchmark datasets covered three sources
of unwanted variation: assay-associated variation in the Perirhinal cortex and
Renal datasets, donor-associated variation in the Liver dataset, and
study/sample-level batch variation in the COVID-19 dataset.

For each pretrained model, only the encoder-side representation modules were
used and all model parameters were frozen. No fine-tuning, adversarial batch
correction or downstream optimization was performed. Input preprocessing
followed the corresponding pretraining procedure, and the final
\texttt{<cls>} hidden state was used as the cell embedding. Embeddings were
normalized to unit $\ell_2$ norm before evaluation, while cell type and batch
labels were used only for calculating integration metrics.

Biological conservation was evaluated using a 15-nearest-neighbor graph
constructed from the normalized embeddings with Euclidean distance, followed
by Leiden clustering at resolution 1.0
\cite{Traag2019Leiden}. We calculated normalized mutual information (NMI),
adjusted Rand index (ARI) and cell type average silhouette width
(ASW$_{\mathrm{cell}}$), with silhouette values mapped from $[-1,1]$ to
$[0,1]$. These metrics were combined as

\begin{equation}
\mathrm{AvgBIO}
=
\frac{
\mathrm{NMI}
+
\mathrm{clip}(\mathrm{ARI},0,1)
+
\mathrm{ASW}_{\mathrm{cell}}
}{3}.
\end{equation}

Batch mixing was evaluated within individual cell types to avoid rewarding
the artificial mixing of biologically distinct populations. Only cell types
containing at least 30 cells and represented in at least two batches were
included. Within each eligible cell type, batch ASW was defined as

\begin{equation}
\mathrm{ASW}_{\mathrm{batch}}
=
1-
\left|
\mathrm{Silhouette}_{\mathrm{batch}}
\right|,
\end{equation}

and graph connectivity was calculated on the same 15-nearest-neighbor graph
as the fraction of cells contained in the largest connected component of the
corresponding cell type subgraph. Batch ASW and graph connectivity were
macro-averaged across eligible cell types and combined as

\begin{equation}
\mathrm{AvgBATCH}
=
\frac{
\mathrm{ASW}_{\mathrm{batch}}
+
\mathrm{GraphConn}
}{2}.
\end{equation}

Higher AvgBIO and AvgBATCH indicate stronger biological conservation and
batch mixing, respectively. When a single summary metric was required, the
overall integration score was calculated as

\begin{equation}
\mathrm{OverallScore}
=
0.6\times\mathrm{AvgBIO}
+
0.4\times\mathrm{AvgBATCH}.
\end{equation}

UMAP projections were generated from the pretrained cell embeddings for
qualitative visualization and colored according to cell type and batch
annotations \cite{McInnes2018UMAP}.

% =========================================================
% Genetic perturbation response prediction
% =========================================================

\subsection{Genetic perturbation-response prediction}
\label{sec:perturbation_prediction}

Genetic perturbation-response prediction was evaluated using GEARS as a
shared downstream framework
\cite{roohaniPredictingTranscriptionalOutcomes2024}. Contextual gene
embeddings were extracted from frozen scKITE and baseline single-cell
foundation models using control-cell expression profiles. These gene
embeddings replaced the original gene embeddings in GEARS while keeping the
perturbation embeddings, gene co-expression graph, Gene Ontology graph and
prediction architecture unchanged. The original GEARS model with native gene
embeddings was included as a baseline. All models were evaluated using the
same data split and evaluation protocol.

Control-cell expression profiles were used for extracting gene embeddings to
avoid information leakage from perturbed states. The extracted gene
representations were aligned to the GEARS gene space, and only genes shared
between the representation space and perturbation datasets were retained for
downstream prediction.

The GEARS simulation splitting procedure was used with a random seed of 1.
Initially, 75\% of unique perturbation genes were assigned to the seen-gene
set. Single-gene perturbations involving the remaining genes formed the
\emph{unseen single} test group. Double perturbations were assigned to
\emph{combo-seen-0}, \emph{combo-seen-1} or \emph{combo-seen-2} according
to whether zero, one or two constituent genes belonged to the initial
seen-gene set. For combinations of two seen genes, 75\% of conditions were
assigned to the initial training pool and the remainder to testing. A
validation set was subsequently constructed from the initial training pool
using the same splitting procedure with gene and combination retention
fractions of 0.9. Evaluation used the saved split and subgroup assignments.

Each downstream model was trained for 15 epochs. The checkpoint with the
lowest validation MSE on the top 20 differentially expressed (DE) genes was
selected for final testing.

Predicted and observed expression profiles were averaged separately across
cells within each perturbation condition. Mean squared error (MSE) and
Pearson correlation were calculated between predicted and observed
condition-mean expression profiles over genes. Unless otherwise specified,
prediction performance was evaluated using the top 20 DE genes. The
benchmark-provided DE gene lists were used throughout; non-dropout analyses
used the separately defined top-20 non-dropout DE gene lists. Condition-level
metrics were macro-averaged with equal weight per original condition label
for the complete test set and each perturbation subgroup.

Direction-correct recovery of top-20 DE genes was evaluated in the Norman
dataset to assess whether predicted perturbation-responsive genes matched
experimentally observed responses. For each perturbation condition, the
observed and predicted top-20 DE gene sets were compared. A gene was
considered correctly recovered only when it was shared between the two sets
and exhibited the same direction of expression change relative to control.
The recovery score was calculated as the fraction of correctly recovered
genes among the 20 observed DE genes.

Genetic interaction magnitude analysis was performed following the GEARS
interaction-analysis framework. For each double-gene perturbation
\(A+B\), expression changes were calculated relative to control cells:

\[
\delta_A=X_A-X_{ctrl},
\]

\[
\delta_B=X_B-X_{ctrl},
\]

\[
\delta_{AB}=X_{AB}-X_{ctrl}.
\]

The double-perturbation response was modeled using robust Theil--Sen
regression without an intercept:

\[
\delta_{AB}=c_A\delta_A+c_B\delta_B+\epsilon .
\]

The genetic interaction magnitude was defined as the Euclidean norm of the
regression coefficients:

\[
\mathrm{GI\ magnitude}=\sqrt{c_A^2+c_B^2}.
\]

Ground-truth genetic interaction magnitudes were calculated from observed
expression profiles, whereas predicted magnitudes were calculated from model
predictions. The agreement between predicted and ground-truth magnitudes was
assessed using Pearson correlation across held-out double-gene perturbations.

% =========================================================

% =========================================================
% 7. Mechanistic interpretation
% =========================================================

\section{Mechanistic interpretation}
\label{sec:annotation_attention}
\subsection{Annotation decoder cross-attention analysis}
Marker genes were identified from the training set following commonly used single-cell marker identification strategies. For each cell type, one-versus-rest differential-expression analysis was performed using a Wilcoxon rank-sum test \cite{stuart2019comprehensive,wilcoxon1945individualcomparisons}. Genes were filtered based on Benjamini--Hochberg-adjusted significance \cite{Benjamini1995FDR}, log-fold change, and detection frequency, and the top 30 genes were retained as the marker gene set. Expression and detection-frequency statistics used for background matching were calculated exclusively from the training set. Marker gene sets and matching statistics were fixed before evaluation on the validation set.

Decoder-to-encoder cross-attention was quantified in validation cells using the frozen scKITE encoder and annotation decoder without expression masking. For each cell, the cell-type span within the natural-language annotation was identified, and the corresponding decoder-to-encoder cross-attention weights were extracted. For a cell-type span containing $T$ decoder tokens, attention weights were averaged across decoder positions and attention heads to obtain a raw gene-level attention score $s_g$ for each input gene:

\[
s_g=
\frac{1}{TH}
\sum_{t=1}^{T}
\sum_{h=1}^{H}
\alpha_{t,h,g},
\]

where $\alpha_{t,h,g}$ denotes the cross-attention weight from decoder position $t$ and attention head $h$ to gene $g$, and $H$ denotes the number of attention heads.

Raw attention scores were converted into within-cell percentile ranks to enable comparison among genes within each cell. For each gene $g$, the attention percentile was calculated as:

\[
p_g=
1-\frac{\mathrm{rank}(s_g)-1}{N-1},
\]

where $N$ represents the number of input genes in the cell and $\mathrm{rank}(s_g)$ denotes the descending rank of the raw attention score among all input genes. Higher $p_g$ values indicate higher relative prioritization by the annotation decoder.

For each validation cell, marker-gene attention was summarized as the median attention percentile across available marker genes:

\[
A_{\mathrm{marker}}
=
\mathrm{median}
\left(
\{p_g|g\in G_{\mathrm{marker}}\}
\right),
\]

where $G_{\mathrm{marker}}$ denotes the marker gene set identified from the training partition.

Matched background genes were selected from non-marker genes based on expression levels and detection frequencies calculated from the training set. Genes were independently divided into five groups according to mean expression and five groups according to detection frequency, resulting in a $5\times5$ matching grid. For each marker gene, background genes were sampled from the same expression-frequency bin while excluding genes belonging to the marker set. For each background sampling replicate, background attention was summarized as:

\[
A_{\mathrm{background}}^{(i)}
=
\mathrm{median}
\left(
\{p_g|g\in G_{\mathrm{background}}^{(i)}\}
\right),
\]

where $i$ denotes the background sampling replicate. The final background attention score was calculated as the average across 20 matched-background replicates:

\[
A_{\mathrm{background}}
=
\frac{1}{20}
\sum_{i=1}^{20}
A_{\mathrm{background}}^{(i)}.
\]

The attention enrichment score was defined as:

\[
\Delta A
=
A_{\mathrm{marker}}
-
A_{\mathrm{background}}.
\]

Marker and matched-background attention scores were subsequently aggregated across cells belonging to the same cell type. Cell types represented by fewer than five evaluable cells were excluded, and paired marker versus matched-background attention values were compared using a two-sided Wilcoxon signed-rank test.

For complementary biological interpretation, canonical marker genes from 14 representative cell populations were manually curated from CellMarker 2.0 \cite{Hu2023CellMarker2}. These canonical markers were used exclusively for visualization and interpretation of cell-type-specific attention patterns and were not used for defining training-derived marker genes or calculating the attention enrichment score.

\subsection{Regulon decoder cross-attention analysis}
\label{sec:regulon_attention}
To quantify regulatory information encoded in the shared representation, 
Regulon Decoder cross-attention was analyzed using the fixed 
validation set. Active regulons were defined based on regulon activity scores 
calculated during preprocessing. For each transcriptomic profile, three active 
regulons were randomly selected from the detected active regulon set using a 
fixed random seed. Regulon identifiers were mapped to their corresponding 
transcription factors (TFs) using the regulon reference constructed during 
pretraining.

For each selected TF, the Regulon Decoder received only the TF query prefix 
without providing any target gene information:
\[
\texttt{<bos> <task> regulon TF1 <gene\_sep> TF2 <gene\_sep>
TF3 <startofanswer>}.
\]

Each TF token was analyzed independently. Decoder-to-encoder cross-attention 
weights were extracted from the final decoder layer at positions corresponding 
to the queried TF tokens. Attention values were averaged across decoder 
attention heads and the corresponding TF query position to obtain a raw 
gene-level regulatory attention score $s_g$:

\[
s_g=
\frac{1}{H}
\sum_{h=1}^{H}
\alpha_{h,g},
\]

where $\alpha_{h,g}$ denotes the cross-attention weight from attention head 
$h$ to encoder gene $g$, and $H$ represents the number of attention heads.

Because raw attention values are normalized within each query position and 
depend on the distribution of attention across input genes, gene-level 
attention scores were converted into within-profile percentile ranks before 
comparison:

\[
p_g=
1-\frac{\mathrm{rank}(s_g)-1}{N-1},
\]

where $N$ represents the number of input genes in the profile and 
$\mathrm{rank}(s_g)$ denotes the descending rank of gene $g$ according to its 
regulatory attention score. Higher $p_g$ indicates stronger relative 
prioritization of gene $g$ during TF query decoding.

External TF--target regulatory interactions were obtained from CollecTRI 
\cite{MullerDott2023CollecTRI}. These interactions were used exclusively for 
external evaluation and were independent of the regulon annotations used during 
scKITE pretraining. For each queried TF, CollecTRI-supported target genes 
present in the encoder input were defined as external target genes:

\[
G_{\mathrm{target}}
=
G_{\mathrm{CollecTRI}}
\cap
G_{\mathrm{encoder}}.
\]

Genes without CollecTRI-supported interactions for the queried TF were 
considered candidate non-target genes. To reduce potential confounding caused 
by gene abundance and detectability, expression and detection statistics were 
calculated exclusively from the training partition. Genes were 
independently divided into five bins according to mean expression level and 
five bins according to detection frequency, resulting in a $5\times5$ 
expression--detection matching grid. For each external target gene, an equal 
number of background genes were randomly sampled from non-target genes within 
the same expression--detection bin. Genes belonging to the CollecTRI target 
set, the corresponding pretraining regulon target set, and the queried TF gene 
itself were excluded from the background candidate pool. Background sampling 
was performed with replacement when necessary and repeated 20 times to reduce 
variation introduced by random background selection.

For each profile--TF pair, target-gene attention and matched-background 
attention were summarized as the median attention percentile across evaluated 
genes:

\[
A_{\mathrm{target}}
=
\mathrm{median}
\left(
\{p_g|g\in G_{\mathrm{target}}\}
\right),
\]

\[
A_{\mathrm{background}}^{(i)}
=
\mathrm{median}
\left(
\{p_g|g\in G_{\mathrm{background}}^{(i)}\}
\right).
\]

The final matched-background attention score was calculated as:

\[
A_{\mathrm{background}}
=
\frac{1}{20}
\sum_{i=1}^{20}
A_{\mathrm{background}}^{(i)}.
\]

Regulatory attention enrichment was defined as:

\[
\Delta A
=
A_{\mathrm{target}}
-
A_{\mathrm{background}}.
\]

Profile-level enrichment scores were subsequently aggregated at the TF level 
using the median across evaluable profiles:

\[
\Delta A_{\mathrm{TF}}
=
\mathrm{median}
(\Delta A_i).
\]

TFs represented by fewer than 10 evaluable profiles were excluded. External 
target and matched-background attention values were compared across TFs using 
a two-sided paired Wilcoxon signed-rank test.

\subsection{Cell-context-specific regulatory attention}
\label{sec:context_regulatory_attention}
To evaluate whether regulatory attention patterns depended on cellular 
context, profile-level regulatory enrichment scores were aggregated for each 
TF--cell type pair using the median across evaluable cells:

\[
\Delta A_{\mathrm{TF,celltype}}
=
\mathrm{median}
(
\Delta A_i
).
\]

TF--cell type combinations represented by fewer than three evaluable cells 
were excluded. TFs and cell types displayed in the heatmap were selected 
according to data coverage rather than enrichment magnitude. Positive values 
indicate preferential attention toward external TF targets compared with 
matched non-target genes, whereas missing combinations indicate insufficient 
evaluable profiles and were left uncolored rather than assigned zero values.

\end{document}